%% file: main.tex
\documentclass[11pt]{article} 
\usepackage[T1]{fontenc}
\usepackage{graphicx}
\usepackage{booktabs}
\usepackage[misc]{ifsym}
\usepackage{fancyhdr} 
\usepackage{mwe} 
\usepackage{amsmath}
\usepackage{algpseudocode}
\usepackage{algorithm}
\usepackage{xcolor}
\usepackage{amssymb}
\usepackage{ulem}
\usepackage[margin=1.1in]{geometry} 

\newcommand{\corr}{(\Letter)}

\title{Variable-Agnostic Causal Exploration for Reinforcement Learning}
\author{
    Minh Hoang Nguyen\textsuperscript{1}\corr, Hung Le\textsuperscript{1}, Svetha Venkatesh\textsuperscript{1} \\
    \textsuperscript{1}Applied Artificial Intelligence Institute (A2I2),\\ Deakin University, Australia \\
    \{s223669184, thai.le, svetha.venkatesh\}@deakin.edu.au
}
\date{\today}

\begin{document}
\maketitle


\pagestyle{fancy} 
\pagenumbering{arabic} 

\begin{abstract}
Modern reinforcement learning (RL) struggles to capture real-world cause-and-effect dynamics, leading to inefficient exploration due to extensive trial-and-error actions. While recent efforts to improve agent exploration have leveraged causal discovery, they often make unrealistic assumptions of causal variables in the environments. In this paper, we introduce a novel framework, Variable-Agnostic Causal Exploration for Reinforcement Learning (VACERL), incorporating causal relationships to drive exploration in RL without specifying environmental causal variables. Our approach automatically identifies crucial observation-action steps associated with key variables using attention mechanisms. Subsequently, it constructs the causal graph connecting these steps, which guides the agent towards observation-action pairs with greater causal influence on task completion. This can be leveraged to generate intrinsic rewards or establish a hierarchy of subgoals to enhance exploration efficiency. Experimental results showcase a significant improvement in agent performance in grid-world, 2d games and robotic domains, particularly in scenarios with sparse rewards and noisy actions, such as the notorious Noisy-TV environments.

\end{abstract}


\section{Introduction}

\input{intro.tex}

\section{Related Work}

\input{related.tex}

\section{Methods}

\input{method.tex}

\section{Experiments}

\input{exp.tex}

\section{Conclusion}
This paper introduces VACERL, a framework that enhances RL agent performance
by analyzing causal relationships among agent observations and actions.
Unlike previous methods, VACERL addresses causal discovery without
assuming specified causal variables, making it applicable to variable-agnostic
environments. Understanding these causal relationships becomes crucial
for effective agent exploration, particularly in environments with
complex causal structures or irrelevant actions, such as the Noisy-TV
problem. We propose two methods to leverage the identified causal
structure. Future research could explore other methods utilizing this
structure. Empirical evaluations in sparse reward navigation and robotic
tasks demonstrate the superiority of our approach over baselines.
However, a limitation is the introduction of new hyperparameters,
which require adjustment for different settings. 

\bibliographystyle{splncs04}

\newpage
\renewcommand\thesubsection{\Alph{subsection}}
\input{appendix}

\end{document}

%% file: intro.tex
Reinforcement learning (RL) is a machine learning paradigm wherein
agents learn to improve decision-making over time through trial and
error \cite{sutton2018reinforcement}. While RL has demonstrated remarkable success in environments with dense rewards \cite{mnih2015human,silver2017mastering}, it tends to fail in case of sparse rewards where the agents do not receive feedback for extended periods, resulting in unsuccessful learning.
Such scarcity of rewards is common in real-world problems: e.g., in
a search mission, the reward is only granted upon locating the target.
Prior studies tackle this problem by incentivizing exploration through
intrinsic rewards \cite{tang2017exploration,burda2018exploration},
motivating exploration of the unfamiliar, or with hierarchical reinforcement
learning (HRL) \cite{levy2017learning,zhang2020generating,pitis2020maximum}.
However, these methods encounter difficulties when scaling up to environments
with complex structures as they neglect the causal dynamics of the
environments. Consider the example of a search in two rooms (Fig.
\ref{fig:An-example-of}(a, b)), where the target is in the second
room, accessible only by opening a “door” with a “key” in the first
room. Traditional exploration methods might force the agent to explore
all corners of the first room, even though only the “key” and “door”
areas are crucial. Knowing that the action \textquotedbl pick up
key\textquotedbl{} is the cause of the effect \textquotedbl door
opened\textquotedbl{} will prevent the agent from aimlessly wandering
around the door before the key is acquired. Another challenge with
these approaches is the Noisy-TV problem \cite{burda2018exploration},
where the agent excessively explores unfamiliar states and actions
that may not contribute to the ultimate task. These inefficiencies
raise a new question:\textit{ Can agents effectively capture causality
to efficiently explore environments with sparse rewards and distracting
actions?}

Inspired by human reasoning, where understanding the relationship
between the environmental variables (EVs) helps exploration, causal
reinforcement learning (CRL) is grounded in causal inference \cite{zeng2023survey}.
CRL research often involves two phases: (i) causal structure discovery
and (ii) integrating causal knowledge with policy training \cite{zeng2023survey}.
Recent studies have demonstrated that such knowledge significantly
improves the sample efficiency of agent training \cite{hu2022causality,seitzer2021causal,zhang2020deep}.
However, \textit{current approaches often assume assess to all environmental
causal variables and pre-factorized environments} \cite{seitzer2021causal,hu2022causality},
simplifying the causal discovery phase. In reality, causal variables
are not given from observations, and constructing a causal graph for
all observations becomes a non-trivial task due to the computational
expense associated with measuring causality. Identifying EVs crucial
for downstream tasks becomes a challenging task, thereby limiting
the effectiveness of CRL methods. These necessitate the identification
of a subset of crucial EVs before discovering causality.

This paper introduces the \textbf{\textit{Variable-Agnostic Causal
Exploration for Reinforcement Learning}} (\textbf{VACERL}) framework
to address these limitations. The framework is an iterative process
consisting of three phases: \textit{``Crucial Step Detection}'',
\textit{``Causal Structure Discovery}'', and \textit{``Agent Training
with Causal Information}''. The first phase aims to discover a set
of crucial observation-action steps, denoted as the \textit{$S_{COAS}$}.
The term ``crucial observation-action step'' refers to an observation
and an action pair stored in the agent's memory identified as crucial
for constructing the causal graph. We extend the idea of detecting
crucial EVs to detecting crucial observation-action steps, motivated
by two reasons. Firstly, variables in the environment are associated
with the observations, e.g., the variable ``key'' corresponds to
the agent's observation of the ``key''. Secondly, actions also contribute
to causality, e.g., the agent cannot use the ``key'' without picking
it up. One way of determining crucial observation-action steps involves
providing the agent with a mechanism to evaluate them based on their
contribution to a meaningful task \cite{hung2019optimizing}. We implement
this mechanism using a Transformer architecture, whose task is to
predict the observation-action step leading to the goal given past
steps. We rank the significance of observation-action steps based
on their attention scores \cite{vaswani2017attention} and pick out
the top-ranking candidates since the Transformer must attend to important
steps to predict correctly. 

In Phase 2, we adapt causal structure learning \cite{ke2019learning}
to discover the causal relationships among the observation-action
steps identified in the discovered $S_{COAS}$ set, forming a causal
graph $G$. The steps serve as the nodes of the causal graph, while
the edges can be identified through a two-phase iterative optimization
of the functional and structural parameters, representing the Structure
Causal Model (SCM). In Phase 3, we train the RL agent based on the
causal graph $G$. To prove the versatility of our approach in improving
the sample efficiency of RL agents, we propose two methods to utilize
the causal graph: (i) formulate intrinsic reward-shaping equations
grounded on the captured causal relationship; (ii) treat the nodes
in the causal graph as subgoals for HRL. During subsequent training,
the updated agent interact with the environments, collecting new trajectories
for the agent memory used in the next iteration of Phase 1. 

In our experiments, we use causally structured grid-world and robotic
environments to empirically evaluate the performance improvement of
RL agents when employing the two approaches in Phase 3. This improvement
extends not only to scenarios with sparse rewards but also to those
influenced by the Noisy-TV problem. We also investigate the contributions
of the core components of VACERL, analyzing the emerging learning
behaviour that illustrates the captured causality of the agents. Our
main contributions can be summarized as:
\begin{itemize}
\item We present a novel \textbf{VACERL} framework, which autonomously uncovers
causal relationships in RL environments without assuming environmental
variables or factorized environments. 
\item We propose two methods to integrate our framework into common RL algorithms
using intrinsic reward and hierarchical RL, enhancing exploration
efficiency and explaining agent behaviour.
\item We create causally structured environments, with and without Noisy-TV,
to evaluate RL agents' exploration capabilities, demonstrating the
effectiveness of our approach through extensive experiments.
\end{itemize}

%% file: related.tex
\begin{figure}[t]
\begin{centering}
\includegraphics[width=\linewidth]{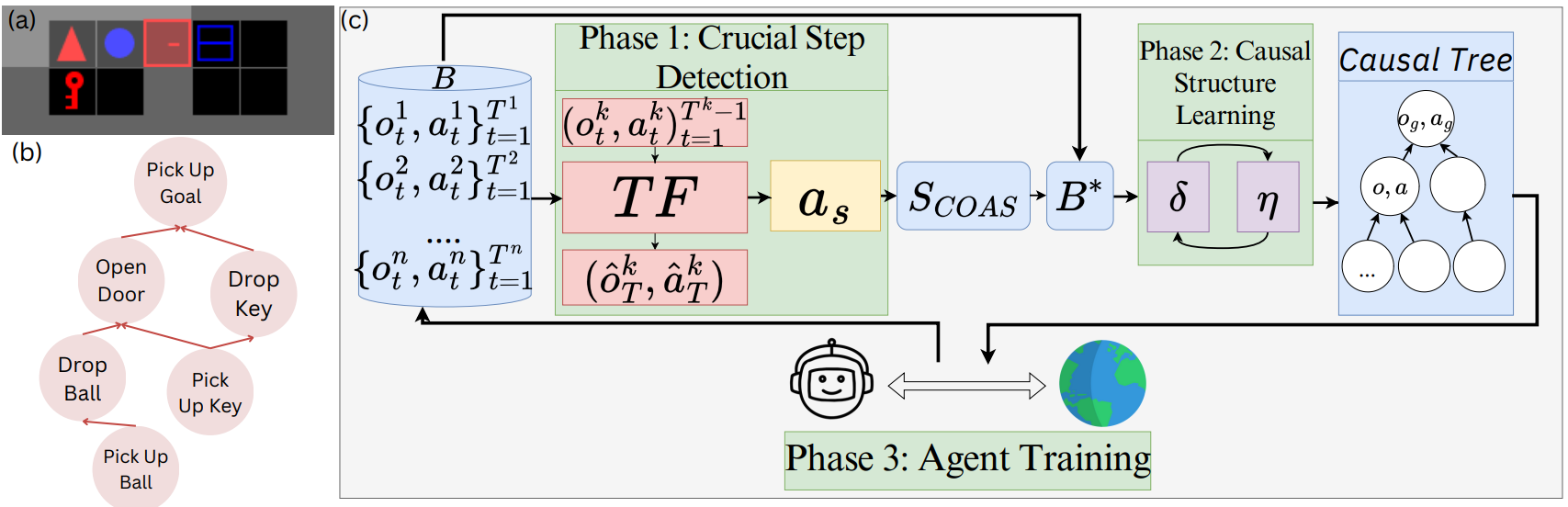}
\par\end{centering}
\caption{\textbf{(a)}: A causally structured environment (MG-2): the agent,
starting in the left room is given a +1 reward when it picks up the
blue box located in the right room. \textbf{(b)}: A possible causal
the graph represents the ideal causality steps for the environment in
(a): pick up and then drop the ball in another position; pick up the
key to open the door; and ultimately pick up the target goal. \textbf{(c)}:
VACERL framework. During the process of the agent training (initially,
a random policy is used) in the environment, we extracted any successful
trajectories and filled buffer $B$. The Transformer ($TF$) model,
trained using $B$, takes input $\{\ensuremath{o_{t}^{k},a_{t}^{k}\}_{t=1}^{T^{k}-1}}$to
predict\textcolor{red}{{} }$(\hat{o}_{T}^{k},\hat{a}_{T}^{k})$. $TF$'s
attention score ($a_{s}$) is used to determine the set $S_{COAS}$
and buffer $B^{*}$. Parameters $\delta$ and $\eta$ of the $SCM$
are optimized using $B^{*}$. We extract a causal hierarchy tree from
the causal graph and use it to design the approaches employed for
agent training. \label{fig:An-example-of}}
\end{figure}

\textbf{Causal Reinforcement Learning} (CRL) is an emerging field
that integrates causality and reinforcement learning (RL) to enhance
decision-making in RL agents, addressing limitations associated with
traditional RL, such as sample efficiency and explainability \cite{zeng2023survey}.
CRL methods can be categorized based on their experimental setups,
whether they are online or offline \cite{zeng2023survey}. Online-CRL
involves real-time interaction with the environment \cite{corcoll2020disentangling,hu2022causality,zhang2020deep,seitzer2021causal},
while Offline-CRL relies on learning from a fixed previously collected
dataset \cite{sun2023offline,pitis2020counterfactual}. Our framework
operates online, using trajectories from an online policy for an agent
training while simultaneously constructing the underlying causal graph.
Prior works in CRL have focused on integrating causal knowledge into
RL algorithms and building causal graphs within the environment. Pitis
et al., \cite{pitis2020counterfactual} use Transformer model attention
weights to generate counterfactual data for training RL agents, while
Coroll et al.,~\cite{corcoll2020disentangling} use causal effect
measurement to build a hierarchy of controllable effects. Zhang et
al., \cite{zhang2024interpretable} measure the causal relationship
between states and actions with the rewards and redistribute the rewards
accordingly. For exploration purposes, CRL research integrates causal
knowledge by rewarding the agents when they visit states with higher
causal influence \cite{seitzer2021causal,zhang2020deep} or treating
the nodes of the causal graph as potential subgoals in HRL \cite{hu2022causality}.
Zhang et al.,~\cite{zhang2020deep} measure the average causal effect
between a predefined group of variables and use this as a reward signal,
meanwhile, Seitzer et al.,~\cite{seitzer2021causal} propose conditional
mutual information as a measurement of causal influence and use it
to enhance the exploration of the RL agent. Hu et al.,~\cite{hu2022causality}
introduce a continuous optimization framework, building a causal structure
through a causality-guided intervention and using it to define hierarchical
subgoals. Despite advancements, previous methods often assume prior
knowledge of EVs and the ability to factorize the environment accordingly.
Our framework autonomously detects crucial steps associated with the key
EVs, enabling causal structure learning without predefined EVs, thus,
distinguishing it from previous methods. The causal graph uncovered
by VACERL is versatile and can complement existing RL exploration
methods, such as intrinsic reward motivation or as hierarchical subgoals. 

\textbf{Intrinsic Reward Motivation}\textbf{\textcolor{red}{{} }}addresses
inefficient training in sparse reward RL environments; an issue associated
with random exploration techniques like $\epsilon$-greedy \cite{bellemare2016unifying}.
The core idea underlying these motivation strategies is to incorporate
intrinsic rewards, which entail adding bonuses to the environment
rewards to facilitate exploration \cite{bellemare2016unifying,tang2017exploration,burda2018exploration}.
These methods add bonuses to environment rewards to encourage exploration,
either based on prediction error \cite{burda2018exploration} or count-based
criteria \cite{bellemare2016unifying,tang2017exploration}. However,
they struggle to scale to complex structure environments, especially
in the scenario of Noisy-TV, where the agent becomes excessively curious
about unpredictable states and ignores the main task \cite{burda2018exploration}.
VACERL tackles this by incorporating a mechanism to identify essential
steps for the primary task and construct the causal graph around these
steps, thus, enabling the agent to ignore actions generating noisy-TV. 

\textbf{Goal-conditioned Hierarchical Reinforcement Learning (HRL)}
is another approach that is used to guide agent exploration. Levy
et al., \cite{levy2017learning} propose a multilevel policies framework,
in which each policy is trained independently and the output of higher-ranking
policies are used as subgoals for lower-level policies. Zhang et al.,
\cite{zhang2020generating} propose an adjacency constraint method
to restrict the search space of subgoals, whereas, Pitis et al., \cite{pitis2020maximum}
introduce a method based on maximum entropy gain motivating the agent
to pursue past achieved goals in sparsely explored areas. However,
traditional HRL methods often rely on random subgoals exploration,
which has shown inefficiency in learning high-quality hierarchical
structures compared to causality-driven approaches \cite{hu2022causality,ding2022generalizing}.
Hu et al.,~\cite{hu2022causality} operate under the assumption of
pre-availability and disentanglement of causal EVs from observations,
using these EVs as suitable subgoals for HRL. However, they overlook
cases where these assumptions are not applicable, e.g., the observation
is the image. In our apprroach, subgoals are determined by abstract representations
of the observation and action, thereby, extending the applications
of causal HRL to unfactorized environments.

%% file: method.tex
\subsection{Background}

\subsubsection*{RL Preliminaries. }

We are concerned with the Partially Observable Markov Decision Process
(POMDP) framework, denoted as the tuple $(S,A,O,P,Z,r,\gamma)$. The
framework includes sets of states $S,$ actions $A$, observations
$O$ providing partial information of the true state, a transition
probability function $P(s'\mid s,a)$, and an observation model $Z$
denoted as $Z(o\mid s,a)$, indicating the probability of observing
$o$ when taking action $a$ in state $s$. $r:S\times A\rightarrow R$
is a reward function that defines the immediate reward that the agent receives
for taking an action in a given state, and discount factor $\gamma$.
The objective of the RL agent is to maximize the expected discounted
cumulative reward $E_{\pi,P}\left[\sum_{t=0}^{\infty}\gamma^{t}r(s_{t},a_{t})\right]$,
over a policy function $\pi$ mapping a state to a distribution over
actions.

\subsubsection*{Causality.}

Causality is explored through the analysis of relationships among
variables and events \cite{pearl2009causal}. It can be described
using the SCM framework \cite{pearl2009causal}. SCM, for a finite
set $V$ comprising $M$ variables, is $V_{i}:=f_{i}(\mathrm{PA}(V_{i})_{(G)},U_{i}),\forall i\in\{1,\ldots,M\},$
where $F=\{f_{1},f_{2},...,f_{M}\}$ denotes the set of generating
functions based on the causal graph $G$ and $U=\{U_{1},U_{2},...,U_{M}\}$
represents the set of noise in the model. The graph $G=\{V,E\}$ provides
the edge $e_{ij}\in E$, representing variable $V_{i}$ causes on
variable $V_{j}$, where $e_{ij}=1$ if $V_{j}\in\mathrm{PA}(V_{i})$,
else, $e_{ij}=0$. The SCM framework can be characterized by two parameter
sets: the functional parameter $\delta$, representing the generating
function $f$; the structural parameter $\eta\in R^{M\times M}$,
modelling the adjacency matrix of $G$ \cite{ke2019learning}.

\subsection{Variable-Agnostic Causal Exploration Reinforcement Learning Framework }

\subsubsection{Overview.}

The primary argument of VACERL revolves around the existence of a
finite set of environment variables (EVs) that the agent should prioritize
when constructing the causal graph. We provide a mechanism to detect
these variables, aiming to reduce the number of nodes in the causal
graph mitigating the complexity of causal discovery. Initially, we
deploy an agent to randomly explore the environment and gather successful
trajectories. Once the agent accidentally reaches the goal a few times,
we initiate Phase 1, reformulating EVs detection into finding the
\textit{``crucial observation-action steps''} ($COAS$) from the
collected trajectories. The agent is equipped with the ability to
rank the importance of these steps by employing the Transformer ($TF$)
model's attention scores ($a_{s}$). Top-$M$ highest-score steps
will form the crucial set $S_{COAS}$. Subsequently, in Phase 2, we
identify the causal relationships among steps in $S_{COAS}$ to learn
the causal graphs $G$ of the environment. In Phase 3, where we extract
a hierarchy causal tree from graph $G$ and use it to design two approaches,
enhancing the RL agent's exploration capability. We then utilize the
updated agent to gather more successful trajectories and repeat the
process from Phase 1. See Fig. \ref{fig:An-example-of}(c) for an
overview of VACERL and detailed implementation in Supp. A 
\footnote{The source is available at https://github.com/mhngu23/Variable-Agnostic-Causal-Exploration-for-Reinforcement-Learning-VACERL}. 

\subsubsection{Phase 1: Crucial Step Detection.}

We hypothesize that important steps (a step is a pair of observation
and action) are those in the agent's memory that the agent must have experienced
to reach the goal. Hence, these steps should be found in trajectories
where the agent successfully reaches the goal. We collect a buffer
$B=(\{o_{t}^{1},a_{t}^{1}\}_{t=1}^{T^{1}},\{o_{t}^{2},a_{t}^{2}\}_{t=1}^{T^{2}},\ldots,\{o_{t}^{n},a_{t}^{n}\}_{t=1}^{T^{n}})$,
where \textit{$n$} is the number of episodes wherein the agent successfully
reaches the goal state, $o_{t}$ and $a_{t}$ is the observation and
action, respectively at step $t$ in an episode, and $T^{k}$ is the
number of steps in the $k$-th episode. We train the $TF$ model,
whose input consists of steps from the beginning to the second-to-last
step in each episode and the output is the last step. The reasoning behind choosing the last step as the prediction target is that it highlights which steps in the trajectories are crucial for successfully reaching the goal.  For a training
episode $k$-th sampled from $B,$ we predict $(\hat{o}_{T}^{k},\hat{a}_{T}^{k})=TF(\{\ensuremath{o_{t}^{k},a_{t}^{k}\}_{t=1}^{T^{k}-1})}$.
The model is trained to minimize the loss $\mathcal{L}_{TF}=\text{\ensuremath{\mathbb{E}_{k}}}\left[MSE\left((\ensuremath{o_{T}^{k},a_{T}^{k}),\ensuremath{(\hat{o}_{T}^{k},\hat{a}_{T}^{k})}}\right)\right]$,
where $MSE$ is the mean square error. Following training, we rank
the significant observation-action steps based on their attention
scores $a_{s}$ (detailed in Supp. A) 

and pick out the top-$M$ highest-score steps. We argue that the top-attended
steps should cover crucial observations and actions that contribute
to the last step prediction task, associated with meaningful causal
variables. For instance, observing the key and the action of picking
it up are linked to the variable ``key''. 

In continuous state space, the agent may repeatedly attend to similar
steps involving the same variable. For example, the agent might select
multiple instances of observing the key, from different positions
where the agent is located, and picks it up. As a result, the set
$S_{COAS}$ will be filled with similar steps relating to picking
up the key and ignoring other important steps. To address this, we
introduce a function $\mathtt{is\_sim}$ to decide if two steps are
the same: 

\begin{itemize}
    \item For discrete action space environments, \(\mathtt{is\_sim}\left(\left(o,a\right),\left(o',a'\right)\right) = 1\)\\ if \(\cos\left(o, o'\right)>\phi_{\text{sim}}\) and \(a = a'\), else \(0\).
    \item For continuous action space environments, \(\mathtt{is\_sim}\left(\left(o,a\right),\left(o',a'\right)\right)=1\) if \(\cos\left((o,a),(o',a')\right)>\phi_{\text{sim}}\), else \(0\).
\end{itemize}
where $Cos\left(o,o'\right)=\frac{o\cdot o'}{\|o\|\cdot\|o'\|}$
and $\phi_{sim}$ is a similarity threshold. Intuitively, if the agent
has two observations with a high cosine similarity and takes the same
action, these instances are grouped. The score $a_{s}$ for a group
is the highest $a_{s}$ among the steps in this group. The proposed $\mathtt{is\_sim}$ method will also be effective in noisy environments, particularly when the observations are trained representations rather than raw pixel data. Subsequently,
we add the steps with the highest $a_{s}$ to $S_{COAS}$. We define
an abstract function $\mathcal{I}$ to map a pair $\left(o_{t}^{k},a_{t}^{k}\right)$
to an element $i$ in $S_{COAS}$: $i=\mathcal{I}\left(\left(o_{t}^{k},a_{t}^{k}\right)\right)\iff\mathtt{is\_sim}\left(\left(o_{t}^{k},a_{t}^{k}\right),\left(o,a\right)_{i}\right)=1$
and collect a new buffer $B^{*}$, where: 

\begin{equation}
B^{*}=B\backslash\left\{ \left(o_{t}^{k},a_{t}^{k}\right):\mathcal{I}\left(\left(o_{t}^{k},a_{t}^{k}\right)\right)\notin S_{COAS}\right\} \label{eq:buffer*}
\end{equation}
Here, $B^{*}$ is $B$ removing steps that are unimportant (not in
$S_{COAS}$).

\subsubsection{Phase 2: Causal Structure Discovery.}

Inspired by the causal learning method proposed by Ke et al.,~\cite{ke2019learning},
we uncover the causal relationships among $M$ steps identified in
the $S_{COAS}$ set. Our approach optimizes the functional parameter
$\delta$ and the structural parameter $\eta$ associated with the
SCM framework. The optimization of these parameters follows a two-phase
iterative update process, wherein one parameter is fixed while the
other is updated. Both sets of parameters are initialized randomly
and undergo training using the buffer $B^{*}$ (Eq. \ref{eq:buffer*}).
Our intuition for training the SCM is that the ``cause'' step has
to precede its ``effect'' step. Therefore, we train the model to
predict the step at timestep $t$ using the sequence of steps leading
to that particular timestep.

In the first causal discovery phase, we fix $\eta$ and optimize $\delta$.
For a step $t$ in the trajectory $k$-th, we formulate $f$ as:

\begin{equation}
(\hat{o}_{t}^{k},\hat{a}_{t}^{k})=f_{\delta,\mathcal{I}\left((o_{t}^{k},a_{t}^{k})\right)}(\{o_{t'}^{k},a_{t'}^{k}\}_{t'=1}^{t-1}\wedge\text{\ensuremath{\left(\mathcal{I}(o_{t'}^{k},a_{t'}^{k})\right)}\ensuremath{\ensuremath{\in}\ensuremath{\mathrm{PA}\left(\mathcal{I}(o_{t}^{k},a_{t}^{k})\right)}}}|G)\label{eq:f_formula}
\end{equation}
 where $\{o_{t'}^{k},a_{t'}^{k}\}_{t'=1}^{t-1}$ is the sequence of
steps from $1$ to $t-1$ that belong to the parental set of $\mathrm{PA}\left(\mathcal{I}(o_{t}^{k},a_{t}^{k})\right)$,
as defined by the current state of $G$ parameterized by $\eta$.
We use MSE as the loss function:

\begin{equation}
\mathcal{L}_{\delta,G}=\mathbb{E}_{t,k}\left[MSE\left((o_{t}^{k},a_{t}^{k}),(\hat{o}_{t}^{k},\hat{a}_{t}^{k})\right)\right]\label{eq:MSE_causal}
\end{equation}

In the second phase, we fix $\delta$ and optimize the parameter $\eta$
by updating the causality from variable $X_{j}$ to $X_{i}$ as $\eta_{ij}=\eta_{ij}-\text{\ensuremath{\beta}}\sum_{h}\left((\sigma(\eta_{ij})-e_{ij}^{(h)})\mathcal{L}_{\delta,G^{(h)},i}(X)\right)$,
where $h$ indicates the $h$-th drawn sample of causal graph $G$,
given the current parameter $\eta$, and $\beta$ is the update rate.
$e_{ij}^{(h)}$ is the edge from variable $X_{j}$ to $X_{i}$ of
$G^{(h)}$, and $\sigma$ is the sigmoid function. $\mathcal{L}_{\delta,G^{(h)},i}(X)$
is the MSE loss in Eq. \ref{eq:MSE_causal} for specific variable
$X_{i}$ of current function $f_{\delta,X_{i}}$ under graph $G^{(h)}$.
After updating parameter $\eta$ for a number of steps, we repeat
the optimization process of parameter $\delta$. Finally, we use the
resulting structural parameter $\eta$ to construct the causal graph
$G$. We derive edge $e_{ij}$ of graph $G$, using: 

\begin{equation}
e_{ij}=\begin{cases}
1 & \text{if }\eta_{ij}>\eta{}_{ji}\text{ and }\sigma(\eta_{ij})>\phi_{causal}\\
0 & \text{otherwise}
\end{cases}\label{eq:edge_formula}
\end{equation}
where $\phi_{causal}$ is the causal threshold.

\subsubsection{Phase 3: Agent Training with Causal Information.}

We extract a refined hierarchy causal tree from graph $G$ with an
intuition to focus on steps that are relevant to achieving the goal.
Using the goal-reaching step as the root node of the tree, we recursively
determine the parental steps of this root node within graph $G$,
and subsequently for all identified parental steps. This causal tree
is used to design causal exploration approaches. These approaches
include (i) intrinsic rewards based on the causal tree, and (ii) utilizing
causal nodes as subgoals for HRL. For the first approach, we devise
a reward function where nodes closer to the root are deemed more important
and receive higher rewards, preserving the significance of the reward
associated with the root node and maintaining the agent's focus on
the goal. In the second approach, subgoals are sampled from nodes
in the causal tree, with nodes closer to the root sampled more frequently.
We present the detailed implementations and empirically evaluate these
approaches in Sec. \ref{subsec:Exploration with Causal Intrinsic Reward}
and Sec. \ref{subsec:Causal subgoals for HRL}.

%% file: exp.tex
\subsection{VACERL: Causal Intrinsic Rewards - Implementation and Evaluation
\label{subsec:Exploration with Causal Intrinsic Reward}}

\subsubsection*{Causal Intrinsic Reward.}

To establish the relationship where nodes closer to the goal hold
greater importance, while ensuring the agent remains focused on the
goal, we introduce intrinsic reward formulas as follows:

\begin{align}
r_{causal}\left(o,a\right) & =r_{g}-\left(d-1\right)r_{0}~\forall(o,a)\in D_{d}\label{eq:reward_1}
\end{align}
where $r_{g}$ is the reward given when the agent reach the goal,
$r_{causal}\left(o,a\right)$ is the intrinsic reward given to a node
$(o,a)$, $D_{d}$ is the set of nodes at depth $d$ of the tree,
$r_{0}=\alpha(r{}_{g}/h)$ with $\alpha$ is a hyperparameter and
$h$ is the tree height. In the early learning stage, especially for
hard exploration environments, the causal graph may not be well defined
and thus, $r_{causal}$ may not provide a good incentive. To mitigate
this issue, we augment $r_{causal}$ with a count-based intrinsic
reward, aiming to accelerate the early exploration stage. Intuitively,
the agent is encouraged to visit never-seen-before observation-action
pairs in early exploration. Notably, unlike prior count-based methods
\cite{bellemare2016unifying}, we restrict counting to steps in $S_{COAS}$,
i.e., only crucial steps are counted. Our final intrinsic reward is:

\begin{equation}
r_{causal}^{+}=\left(1/\sqrt{n_{(o,a)_{t}}}\right)r_{causal}(o_{t},a_{t})\label{eq:reward_3}
\end{equation}
where $n_{(o,a)_{t}}$ is the number of time observation $o$ and
action $a$ is encountered up to time step $t$. Starting from zero,
this value increments with each subsequent encounter. We add the final
intrinsic reward to the environment reward to train the policy. The
total reward is $r\left(s_{t},a_{t}\right)=r_{env}\left(s_{t},a_{t}\right)+r_{causal}^{+}\left(s_{t},a_{t}\right)$,
where $r_{env}$ is the extrinsic reward provided by the environment. 

\subsubsection*{Environments.}

\noindent We perform experiments across three sets of environments:
FrozenLake (FL), Minihack (MH), and Minigrid (MG). These environments
are tailored to evaluate the approach in sparse reward settings, where
the agent receives a solitary +1 reward upon achieving the goal (detailed
in Supp. B)

\emph{\uline{FL}} includes the 4x4 (4x4FL) and 8x8 (8x8FL) FrozenLake
environments (Supp Fig. B.1(d,e)) 
\cite{towers_gymnasium_2023}.
Although these are classic navigation problems, hidden causal relationships
exist between steps. The pathway of the agent can be conceptualized
as a causal graph, where each node represents the agent's location
cell and its corresponding action. For example, moving right from
the cell on the left side of the lake can be identified as the cause
of the agent falling into the lake cell. We use these environments
to test VACERL's efficiency in discrete state space, where $\mathtt{is\_sim}$
is not used. 

\emph{\uline{MH}} includes MH-1 (Room), MH-2 (Room-Monster), MH-3
(Room-Ultimate) and MH-4 (River-Narrow) \cite{samvelyan2021minihack}.
These environments pose harder exploration challenges compared to
FL due to the presence of more objects. Some environments even require
interaction with these objects to reach the goal, such as killing
monsters (MH-2 and MH-3) or building bridges (MH-4). For this set of environments, we use pixel observations.

\emph{\uline{MG}} is designed based on Minigrid Environment \cite{MinigridMiniworld23},
with escalating causality levels. These include the Key Corridor (MG-1)
(Supp. Fig. B.1(a)) and 3 variants of the BlockUnlockPickUp:
2 2x2 rooms (MG-2 Fig. \ref{fig:An-example-of}(a)), 2 3x3 rooms (MG-3)
and the 3 2x2 rooms (MG-4) (Supp Fig. B.1(b,c)).
The task is to navigate and locate the goal object, in a different
room. These environments operate under POMDP, enabling us to evaluate
the framework's ability to construct the causal graph when only certain
objects are observable at a timestep. In these environments, the agent
completes the task by following the causal steps: firstly, remove
the obstacle blocking the door by picking it up and dropping it in another
position, then, pick up the key matching the colour of the door to
open it; and finally, pick up the blue box located in the rightmost
room, which is the goal. In MG-3, distracting objects are introduced
to distract the agent from this sequence of action. In any case, intrinsic
exploration motivation is important to navigate due to reward sparsity;
however, blind exploration without an understanding of causal relationships
can be ineffective. 

\emph{\uline{Noisy-TV}} setting is implemented as an additional
action (action to watch TV) and can be incorporated into any of the
previous environments, so the agent has the option to watch the TV
at any point while navigating the map \cite{10.5555/3635637.3662964}. When taking this watching TV
action, the agent will be given white noise observations sampled from
a standard normal distribution. As sampled randomly, the number of
noisy observations can be conceptualized as infinite. 

\subsubsection*{Baselines.}

\noindent PPO \cite{schulman2017proximal}, a policy gradient method,
serves as the backbone algorithm of our method and other baselines.
Following Schulman et al.,~\cite{schulman2017proximal}, vanilla PPO
employs a simple entropy-based exploration approach. Other baselines
are categorized into causal and non-causal intrinsic motivation. Although
our focus is causal intrinsic reward, we include non-causal baselines
for comprehensiveness. These include popular methods: Count-based
\cite{bellemare2016unifying,tang2017exploration} and RND \cite{burda2018exploration}.
Causal motivation baselines include ATTENTION and CAI, which are two
methods that have been used to measure causal influence \cite{seitzer2021causal,pitis2020counterfactual}.
We need to adapt these methods to follow our assumption of not knowing
causal variables. The number of steps used to collect initial successful
trajectories and to reconstruct the causal graph (denoted as $H_{s}$
and $T_{s}$ respectively) for VACERL and causal baselines are provided
for each environment in Supp. D. 
However, not all causal methods can be adapted, and as such, we have
not conducted comparisons with approaches, such as \cite{hu2022causality}.
Additionally, as we do not require demonstrating trajectory from experts,
we do not compare with causal imitation learning methods \cite{de2019causal,sun2023offline}.

\begin{figure*}[t]
\begin{centering}
\includegraphics[scale=0.65]{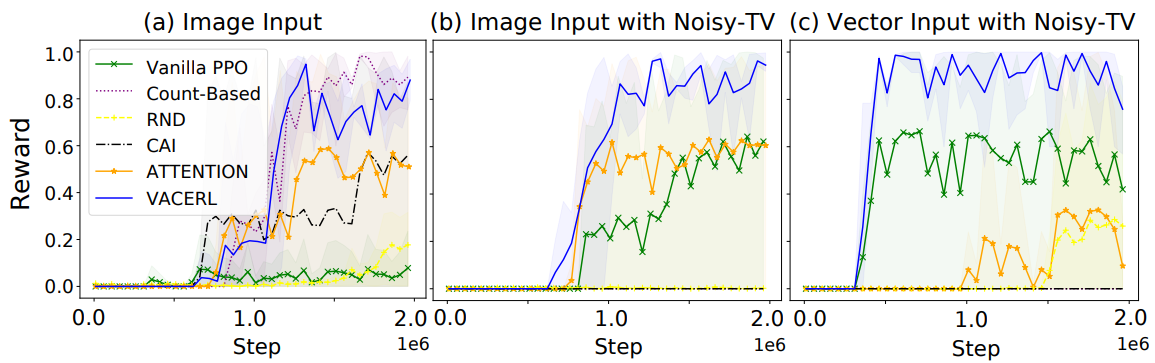}
\par\end{centering}
\caption{The learning curves for MG-2 illustrate the average return (mean$\pm$std
over 3 runs) of 50 testing episodes over 2 million training steps.
For VACERL and causal baselines, the learning curves include 300,000
steps dedicated to the initial random exploration period to collect initial
successful trajectories, with rewards for each step set to 0. The causal
graph is reconstructed every 300,000 steps. \label{fig:ME Results}}
\end{figure*}
\subsubsection*{Results.}
In this section, we present our empirical evaluation results of VACERL with causal intrinsic rewards.

\textit{\uline{Discrete State Space}}: Table \ref{tab:Results} illustrates
that our rewards improve PPO's performance by approximately 30\%,
in both 4x4FL and 8x8FL environments. Notably, VACERL outperforms
both causal baselines, ATTENTION and CAI. Specifically, VACERL surpasses
ATTENTION by 67\% and 39\% in 4x4FL and 8x8FL. CAI fails to learn
the tasks within the specified steps due to insufficient trajectories
in the agent's memory for precise causality estimation between all
steps. In contrast, our method, incorporating a crucial step detection
phase, requires fewer trajectories to capture meaningful causal relationships
in the environment. VACERL also performs better than Count-based by
66\% in 4x4FL and 100\% in 8x8FL, and RND by 51\% in 4x4FL and 31\%
in 8x8FL. We hypothesize that Count-based and RND's intrinsic rewards
are unable to encourage the agent to avoid the trapping lakes, unlike
VACERL's are derived from only successful trajectories promoting safer
exploration. 

\textit{\uline{MG-2 Learning Curve Analysis}}: We conduct experiments
with 2 types of observation space (image and vector) and visualize
the learning curves in Fig. \ref{fig:ME Results}(a) and Supp.
Fig. B.4. Results demonstrate
that VACERL outperforms vanilla PPO, causal baselines, and RND in
both types of observation space. While VACERL shows slightly slower
progress than Count-based in early steps, it quickly catches up in
later stages, ultimately matching optimal performance. We attribute
this to VACERL requires a certain number of training steps to accurately
acquire the causal graph before the resulting causal rewards influence
the agent's training—a phenomenon observed in other causal baselines
as well.

\textit{\uline{Continuous State Space}}: Table \ref{tab:Results}
summarizes the testing results on 8 continuous state space environments
(MH-1 to MG-4). In most of these environments, VACERL demonstrates
superior performance. It only ranks as the second-best in MH-4, MG-1
and MG-2 with competitive returns. In MG-3 environment, at 30 million
steps, VACERL achieves the best result with an average return of 0.77,
outperforming the second-best Count-based by 10\%, while other baselines
show little learning. Notably, in the hardest task MG-4, only VACERL
can show signs of learning, achieving an average score of 0.29 after
50 million steps whereas other baselines' returns remain zero. Additional
learning curves and results are provided in Supp. B.

\textit{\uline{Under Noisy-TV}}: Fig. \ref{fig:ME Results}(b, c),
showing the results on MG-2 environment under Noisy-TV setting, confirm
that our reward exhibits greater robustness in Noisy-TV environments
compared to traditional approaches. Count-based, CAI, and RND fail
in this setting as they cannot differentiate noise from meaningful
novelty, thus, getting stuck watching Noisy-TV. While the noise less impacts ATTENTION and
naive PPO, their exploration strategies
are not sufficient for sparse reward environments. Overall, VACERL
is the only method performing well across all settings, with or without
Noisy-TV. 

\begin{table*}[t]
\centering{}%
\begin{tabular}{cccccccc}
\hline 
Task & Step (,000) & PPO & Count-Based & RND & ATTENTION & CAI & VACERL\tabularnewline
\hline 
4x4FL & 5 & $66\pm47$ & $33\pm57$ & $48\pm50$ & $32\pm56$ & $0\pm0$ & \textbf{$\mathbf{97\pm4}$}\tabularnewline
8x8FL & 35 & $66\pm58$ & $0\pm0$ & $66\pm57$ & $59\pm53$ & $0\pm0$ & \textbf{$\mathbf{96\pm4}$}\tabularnewline
MH-1 & 5 & $66\pm57$ & \textbf{$\mathbf{99\pm0}$} & $88\pm20$ & $66\pm57$ & $33\pm57$ & \textbf{$\mathbf{99\pm0}$}\tabularnewline
MH-2 & 1,000 & $77\pm7$ & $71\pm11$ & $68\pm3$ & $76\pm15$ & $67\pm13$ & \textbf{$\mathbf{85\pm8}$}\tabularnewline
MH-3 & 1,000 & $61\pm8$ & $67\pm3$ & $61\pm8$ & $65\pm3$ & $62\pm7$ & \textbf{$\mathbf{68\pm10}$}\tabularnewline
MH-4 & 5,000 & $42\pm3$ & \textbf{$\mathbf{47\pm7}$} & $42\pm6$ & $42\pm3$ & $45\pm4$ & $46\pm3$\tabularnewline
MG-1 & 500 & $82\pm2$ & $85\pm12$ & $85\pm12$ & \textbf{$\mathbf{98\pm2}$} & $66\pm57$ & $94\pm1$\tabularnewline
MG-2 & 2,000 & $9\pm13$ & \textbf{$\mathbf{94\pm4}$} & $20\pm16$ & $48\pm43$ & $59\pm51$ & $90\pm14$\tabularnewline
MG-3 & 30,000 & $0\pm0$ & $69\pm25$ & $11\pm6$ & $26\pm46$ & $0\pm0$ & \textbf{$\mathbf{77\pm3}$}\tabularnewline
MG-4 & 50,000 & $0\pm0$ & $0\pm0$ & $0\pm0$ & $0\pm0$ & $0\pm0$ & \textbf{$\mathbf{29\pm50}$}\tabularnewline
\hline 
\end{tabular}\caption{FL, MH and MG: Average return ($\times100$) of 50 episodes (mean$\pm$std.
over 3 runs). Bold denotes the best results. For VACERL and causal
baselines (CAI and ATTENTION). Steps include the initial steps, as
detailed in Supp. D,
dedicated to collecting initial successful trajectories. \label{tab:Results}}
\end{table*}

\subsection{VACERL: Causal Subgoals - Implementation and Evaluation \label{subsec:Causal subgoals for HRL}}

\subsubsection*{Causal subgoals sampling.}

In HRL, identifying subgoals often relies on random exploration \cite{levy2017learning,zhang2020generating},
which can be inefficient in large search spaces. We propose leveraging
causal nodes as subgoals, allowing agents to actively pursue these
significant nodes. To incorporate causal subgoals into exploration,
we suggest substituting a portion of the random sampling with causal
subgoal sampling. Specifically, in the HRL method under experimentation
where subgoals are randomly sampled 20\% of the time, we replace a
fraction of this 20\% with a node from the causal tree as a subgoal,
while retaining random subgoals for the remainder. Eq. \ref{eq:probability_nodes}
denotes the probability of sampling a node $i$ at depth $d>0$ (excluding
the root node as this is the ultimate goal) from the causal tree: 

\begin{equation}
P^{(i)}=(d_{i})^{-1}/\sum\limits _{i=1}^{N}(d_{i})^{-1}\label{eq:probability_nodes}
\end{equation}

\noindent with $d_{i}$ is the depth of node $i$ and $N$ is the
number of nodes in the causal tree. 

\subsubsection*{Environments. }

\noindent We use FetchReach and FetchPickAndPlace environments from
Gymnasium-Robotics \cite{gymnasium_robotics2023github}. These are
designed to test goal-conditioned RL algorithms. We opt for sparse
rewards settings, in which only a single reward of $0$ is given if
the goal is met, otherwise $-1$ (detailed in Supp. C). 

\subsubsection*{Baselines.}

\noindent HAC \cite{levy2017learning}, a goal-conditioned HRL algorithm,
serves as the backbone and a baseline. HAC is implemented as a three-level
DDPG \cite{lillicrap2015continuous} with Hindsight Experience Replay
(HER) \cite{andrychowicz2017hindsight}, where the top two levels
employ a randomized mechanism for subgoal sampling. We also evaluate
our performance against the standard DDPG+HER algorithm \cite{andrychowicz2017hindsight}
on the FetchPickAndPlace environment, as this is the more challenging
task \cite{plappert2018multi} and for comprehensiveness.

\subsubsection*{Results.}

\begin{figure*}[t]
\centering{}\includegraphics[width=\linewidth]{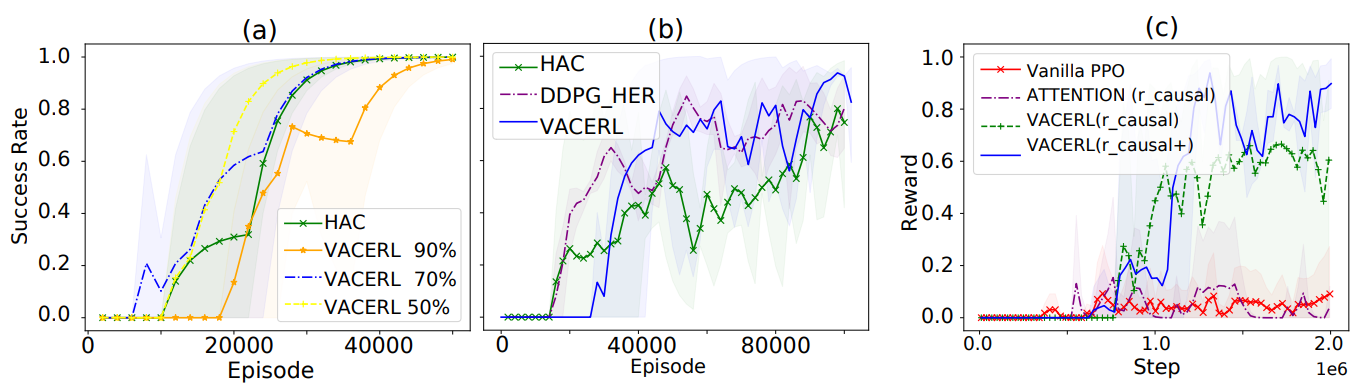}\caption{\textbf{(a, b)}: Learning curves of FetchReach (a) (\% represents
the portion of random subgoals replaced) and FetchPickAndPlace (b).
The average return of 50 testing episodes (mean$\pm$std over 3 runs).
For VACERL, successful trajectories are collected during training
and causal graphs are reconstructed every 2,000 episodes and 10,000
episodes, respectively. \textbf{(c)}: Component contribution study
on MG-2 task. The average return of 50 episodes (mean$\pm$std. over 3
runs). \label{fig:R_env_learning_curve}}
\end{figure*}
In this section, we outline our empirical evaluation results of VACERL with causal subgoals.

\textit{\uline{FetchReach}}: We assess the impact of replacing
varying proportions of random sampling subgoals with nodes from the
causal graph, based on Eq. \ref{eq:probability_nodes}, on the performance
of the HRL agent. The learning curve in Fig. \ref{fig:R_env_learning_curve}(a)
suggests that replacing with percentages of 50\% and 70\% enhances
the sample efficiency of the Vanilla HAC. Notably, when employing
a 70\% substitution rate, agents demonstrate signs of learning after
only 4,000 episodes, a considerable improvement over the HAC agent's
10,000 episodes. Conversely, replacing 50\% leads to a swifter convergence,
at 20,000 episodes comparing to HAC at 25,000 episodes. Additional
experiments (Supp. C)
demonstrate that this accelerated convergence rate is attributable
to the learned causal subgoals. In contrast, employing a 90\% substitution
rate results in a decline in performance. We assert that this decline
comes from insufficient exploration of new subgoals, leading to an
inadequate number of trajectories in buffer $B$ for causal discovery. 

\textit{\uline{FetchPickAndPlace}}: We adopt the 50\% replacement,
which yielded the most stable performance in the FetchReach environment
for this environment. The learning curve of VACERL in Fig. \ref{fig:R_env_learning_curve}(b)
shows a similar pattern to the learning curves for the MG-2 task in
Fig. \ref{fig:ME Results}(a). VACERL progresses slower but eventually
achieves optimal performance, surpassing DDPG+HER and HAC after 90,000
episodes. In this environment, we reconstruct the causal tree every
10,000 episodes, and as seen in the learning curve, the RL agent's
performance begins to improve after approximately 20,000 episodes
(worst case improves after 40,000 episodes). 

\subsection{Ablation Study and Model Analysis }

\begin{figure*}[t]
\centering{}\includegraphics[width=\linewidth]
{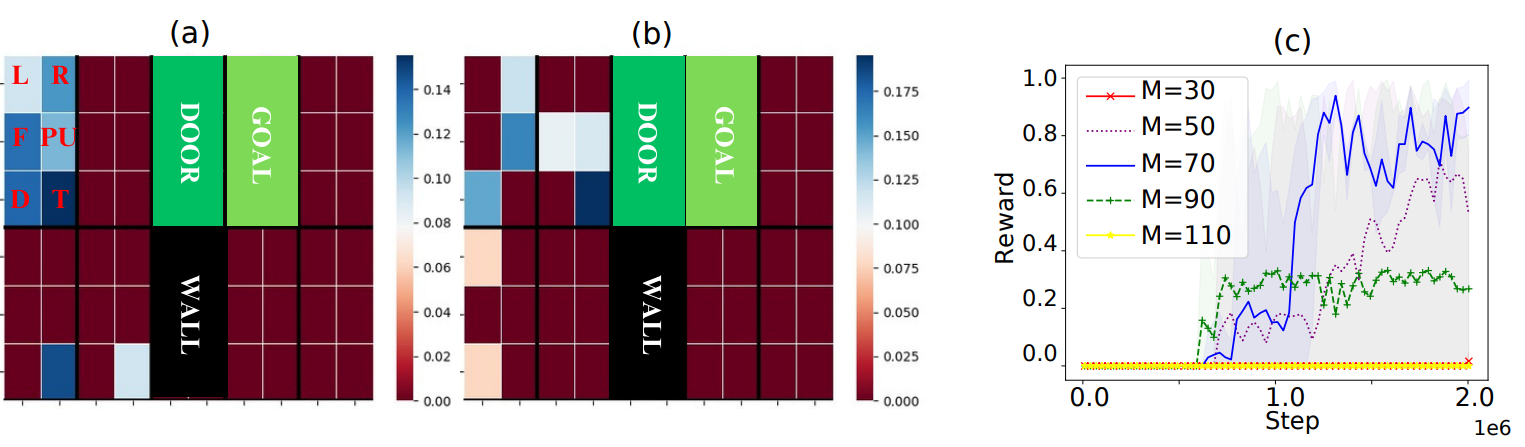}\caption{\textbf{(a,b)}: Attention heatmap when $B$ has 4 (a) and 40 (b) trajectories
for MG-2 task. We highlight the top-8 attended actions and their grids.
A big cell (black boundary) represents a grid in the map, containing
smaller cells representing 6 possible actions. \textbf{(c)}: Results
tuning the number of steps ($M$) in $S_{COAS}$. Average return of
50 episodes (mean$\pm$std. over 3 runs). \label{fig:Ablation_Study}}
\end{figure*}

We use MG-2 (Fig. \ref{fig:An-example-of}(a)) task and causal intrinsic
reward for our analysis.

\textit{\uline{Crucial Step Detection Analysis}}: We investigate
how the the Transformer model $TF$'s performance changes with varying
buffer $B$ sizes\textit{.} As depicted in Fig. \ref{fig:Ablation_Study}(a,b),
increasing the number of trajectories in $B$ enhances the framework's
accuracy in detecting important steps through attention. Initially,
with 4 trajectories (Fig. \ref{fig:Ablation_Study}(a)), $TF$ attends
to all actions in the top-left grid. However, after being trained
with 40 trajectories (Fig. \ref{fig:Ablation_Study}(b)), $TF$ correctly
attends to pick-up action (PU) in the top-left grid, corresponding
to the key pickup event. It can also attend to toggle action (T) in front
of the door, corresponding to using the key to open the door. Additional
visualization for 4x4FL is in Supp. Fig. B.6. 

Next, we investigate the effects of employing varying sizes of $S_{COAS}$
($M$). The results in Fig. \ref{fig:Ablation_Study}(c) reveal that
varying $M$ changes the performance of the agent drastically. If
$M$ is too small, the agent will not be able to capture all causal
relations, thereby failing to mitigate the issue of sparse reward.
On the other hand, too large $M$ can be noisy for the causal discovery
phase as the causal graph will have redundant nodes. We find that
the optimal value for $M$, in MG-2, is 70, striking a balance between
not being too small or too large. 

\textit{\uline{Intrinsic Reward Shaping Analysis}}: We exclude
the counting component (Eq. \ref{eq:reward_3}) from the final intrinsic
reward to assess the agent's exploration ability simply based on $r_{causal}$.
The results in Fig. \ref{fig:R_env_learning_curve}(c) show that the
agent remains proficient relying only on the causally motivated reward
(green curve). In particular, in the absence of Eq. \ref{eq:reward_3},
the VACERL agent still outperforms Vanilla PPO. However, its performance
is not as optimal as the full VACERL ($r_{causal}^{+}$, blue curve).
This is because, in early iteration, the causal graph is not yet well
defined, diminishing the efficiency of solely using the causal intrinsic
reward $r_{causal}$.

\begin{figure}[t]
\centering{}\includegraphics[scale=0.5]{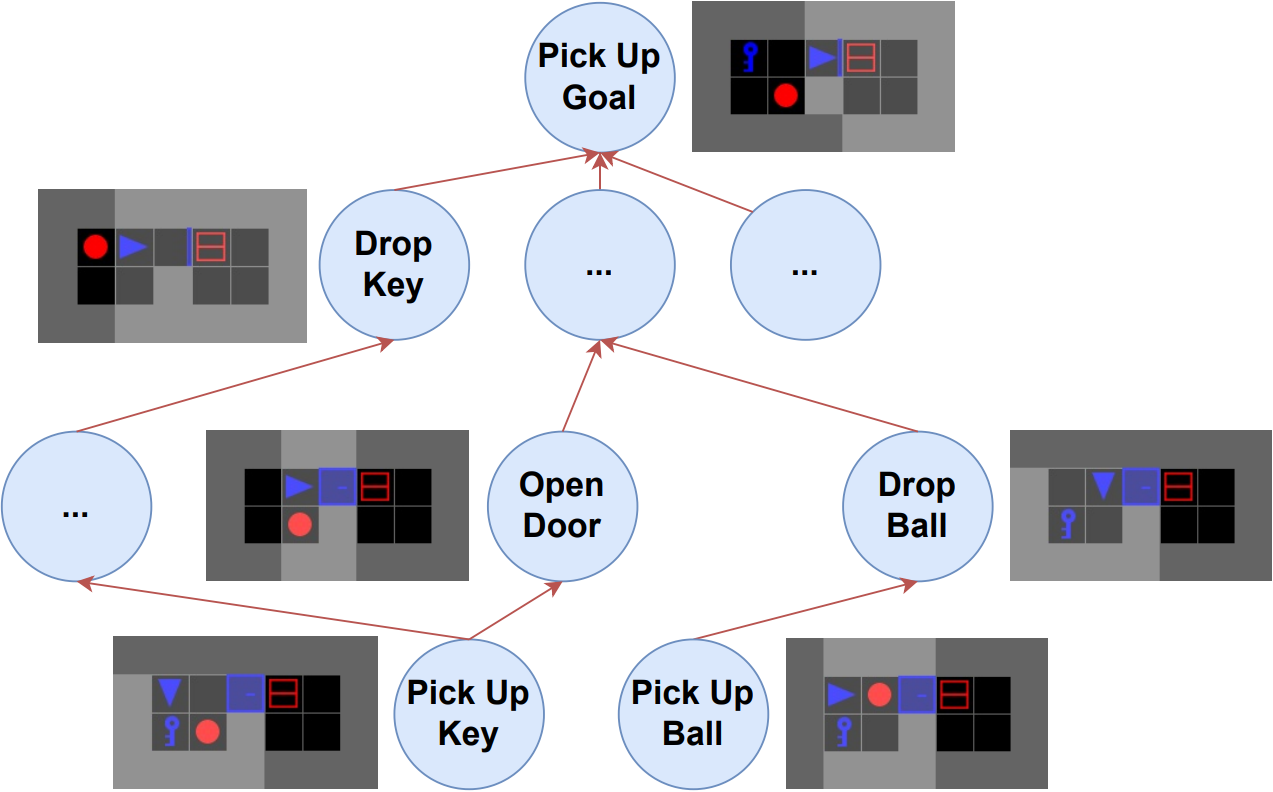}\caption{Causal graph that was generated as a result of the algorithm for MG-2.
The RGB images display the agent's viewports during the execution
of the corresponding actions. \label{fig:generated_causal_graph}}
\end{figure}

\textit{\uline{Causal Structure Discovery Contribution}}: We study
the contribution of Phase 2, comparing between causality and attention
correlation. We directly used the $a_{s}$ assigned to each $(o,a)$
in Phase 1 to compute the intrinsic reward: $r_{bonus}\left(o,a\right)=\alpha a_{s}\left(o,a\right)$,
where $\alpha$ is the hyperparameter in Eq. \ref{eq:reward_1}. This
reward differs from the ATTENTION reward used in Sect. \ref{subsec:Exploration with Causal Intrinsic Reward}
in that the augmentation in Eq. \ref{eq:reward_3} is not applied.
We expect that as attention score is a reliable indicator of correlation,
building intrinsic reward upon it would benefit the agent, albeit
not as effectively as when a causal graph is used (correlation is
not as good as causality). The learning curve in Fig. \ref{fig:R_env_learning_curve}(c)
showcases that the result using causal as intrinsic motivation (green
curve) performs better than using attention correlation (purple curve)
by a large margin. To further evaluate, we extract the
learned causal graph in Fig. \ref{fig:generated_causal_graph} and present the detailed analysis of this graph in Supp. B.
The result shows that our method can recover an approximation of the
ground truth causal graph. Although there are redundant nodes and
edges, important causal hierarchy is maintained, e.g., ``open door''
is the parental step of ``pickup key''.

%% file: appendix.tex
\renewcommand\thesubsection{\Alph{subsection}}

\subsection{Details of Methodology\label{subsec:Details-of-Methodology}}

\setcounter{figure}{0} 

\renewcommand{\thefigure}{\Alph{subsection}.\arabic{figure}} 

\subsubsection{VACERL Framework\label{subsec:VACERL-Framework}\\}

The detailed processing flow of the VACERL framework is described
in Algo. \ref{alg:VACERL-framework.}. Buffer $B$ is initialized
using the process from lines 2-6, using a random policy to collect
successful trajectories (Note: as long as the agent can accidentally
reach the goal and add 1 trajectory to $B$, we can start the improving
process). We, then, start our iterative process (the outer loop).
In Phase 1, ``Crucial Step Detection'' (lines 8-21), the process
commences with the training of the Transformer model $TF$ using Algo.
\ref{alg:train_transformer}. Subsequently, we collect the dictionary
$D$ that maps $(o_{t}^{k},a_{t}^{k})$ to attention score $a_{s}$.
$D$ is sorted based on $a_{s}$. We, then, define $\mathtt{is\_sim}$
function (line 9) and abstract function $\mathcal{I}$ (line 10) to
handle similar observation-action steps, and add the top $M$ $(o_{t}^{k},a_{t}^{k})$
steps to the set $S_{COAS}$ using the process from lines 12-20. After
collecting $S_{COAS}$, we apply Eq. 1 to acquire
the new buffer $B^{*}$. With buffer $B^{*}$, we initiate Phase 2
(line 22) called ``Causal Structure Discovery''. We optimize the
two parameters $\delta$ and $\eta$ using Algo. \ref{alg:train_SCM}
and collect the causal graph $G$. Using graph $G$, we collect the
causal tree relative to the goal-reaching step to create a hierarchy
of steps. We use this hierarchy to calculate the intrinsic reward
associated with $(o,a)$ using Eq. 6 or to calculate
subgoals sampling probability using Eq. 7.
Finally, we train the policy $\pi_{\theta}$ and adding new successful
trajectories to buffer $B$, summarizing Phase 3 (lines 23-29) called
``Agent Training with Causal Information''. The process starts again
from Phase 1 using the updated buffer $B$.

\begin{algorithm}[h]
\begin{algorithmic}[1] 
\label{alg:algo1}
\Require{Environment $env$, policy $\pi_{\theta}$, Buffer $B$, Transformer model $TF$, number crucial steps $M$, crucial step set $S_{COAS}$, and Dictionary $D$}
\State Initialize $\pi_{\theta}$, $B$, $TF$, $S_{COAS}$, $D$
\For {$k = 1,2,...,H_s$}
	\State Execute policy $\pi_{\theta}$ to collect $\{o_{t}^{k},a_{t}^{k}\}_{t=1}^{T^{k}}$  
	\If{$\mathtt{reach\_{goal}}()$} $B \gets \{o_{t}^{k},a_{t}^{k}\}_{t=1}^{T^{k}}$ 
	\EndIf
\EndFor

\For {$iteration = 1,2,...$}
	\State $D$ = $\mathtt{train\_transformer}$($TF$, $B$, $D$) (Algo. \ref{alg:train_transformer})
	\Comment \textbf{Phase 1}
 	\State Define $\mathtt{is\_sim}$ 
	\State Define $\mathcal{I}$, $S_{COAS}: i=\mathcal{I}\left(\left(o_{t}^{k},a_{t}^{k}\right)\right)\iff\mathtt{is\_sim}\left(\left(o_{t}^{k},a_{t}^{k}\right),\left(o,a\right)_{i}\right)=1$
	\State Sort $D$ based on descending order of $D.values$
	\For {$key$ in $D.keys$}:
		\If {$S_{COAS}.length < M$}
			\If {$\mathcal{I}\left(key\right) \notin S_{COAS} $}
					\State $S_{COAS} \gets key$	
			\EndIf
		\Else 
			\State $\mathtt{break}$
		\EndIf
	\EndFor
	\State Collect $B^{*}$ using Eq. 1
	\Comment \textbf{End Phase 1}
	\State $G$ = $\mathtt{train\_SCM}$($B^{*},S_{COAS}, \mathcal{I}, M$) (Algo. \ref{alg:train_SCM})		
	\Comment \textbf{Phase 2}
	\State Extract causal tree from graph $G$ 
	\Comment \textbf{Phase 3}
	\For {$k = 1,2,...,T_s$}
		\State Train policy $\pi_{\theta}$ with $r_{causal}^{+}$ computed using Eq. 6 or with subgoal probability $P^(i)$ using Eq. 7, and collect  $\{o_{t}^{k},a_{t}^{k}\}_{t=1}^{T^{k}}$   
		\If{$\mathtt{reach\_{goal}}()$} $B \gets \{o_{t}^{k},a_{t}^{k}\}_{t=1}^{T^{k}}$ 
		\EndIf
	\EndFor
	\Comment \textbf{End Phase 3}
\EndFor
\end{algorithmic}

\caption{VACERL framework.\label{alg:VACERL-framework.}}
\end{algorithm}

\subsubsection{Transformer Model Training\label{subsec:Transformer-Model-Training}\\}

Detailed pseudocode for training the Transformer model is provided
in Algo. \ref{alg:train_transformer}. We utilize the Transformer
architecture implemented in PyTorch \footnote{{\scriptsize{}https://pytorch.org/docs/stable/generated/torch.nn.Transformer.html}},
for our $TF$ model implementation. This implementation follows the
architecture presented in the paper \cite{vaswani2017attention},
thus, the attention score $a_{s}$ for a step is computed using the
self-attention equation: $\text{softmax}\left(\frac{QK^{T}}{\sqrt{d_{k}}}\right)$,
where $Q=XW_{Q}$ represents the query vector, $K=XW_{K}$ represents
the key vector, $X$ is the learned embedding of a step $(o_{t}^{k},a_{t}^{k})$,
and $W_{Q},W_{K}$ are trainable weights. The values of $a_{s}$ are
extracted from the encoder layer of the $TF$ model during the last
training iteration. We use a step $(o_{t}^{k},a_{t}^{k})$ as the
key in dictionary $D$ that maps to an associated $a_{s}$, described
in the process in lines 5-11 (Algo. \ref{alg:train_transformer}).

\begin{algorithm}[h]
\begin{algorithmic}[1]
\Require{Transformer model $TF$, Buffer $B$ and Dictionary $D$}
\For {$iteration = 1,2,...,I$}
	\For {$k$ episodes in $B$}
		\State $(\hat{o}_{T}^{k},\hat{a}_{T}^{k})=TF(\{\ensuremath{o_{t}^{k},a_{t}^{k}\}_{t=1}^{T^{k}-1})}$
		\State Compute $\mathcal{L}^{k}_{TF}=MSE\left((\ensuremath{o_{T}^{k},a_{T}^{k}),\ensuremath{(\hat{o}_{T}^{k},\hat{a}_{T}^{k})}}\right)$
		\If{$i$==$I$}
		\Comment {$D$ is only updated in the last training iteration}
			\If {$(o_{t}^{k},a_{t}^{k}) \notin D.keys$}
				\State $D[(o_{t}^{k},a_{t}^{k})] \gets a_s$
			\Else
				\State $D[(o_{t}^{k},a_{t}^{k})] \gets max(a_s,D[(o_{t}^{k},a_{t}^{k})])$
			\EndIf
		\EndIf
		\State Update $TF$ by minimizing loss $\mathcal{L}^{k}_{TF}$
	\EndFor
\EndFor
\State \textbf{return} $D$
\end{algorithmic}
\caption{Train Transformer.\label{alg:train_transformer}}
\end{algorithm}

\subsubsection{SCM Training\label{subsec:SCM-Training}\\}

The detailed pseudocode is provided in Algo. \ref{alg:train_SCM}.
Our approach involves a two-phase iterative update process, inspired
by the causal learning method proposed by Ke et al.,~\cite{ke2019learning}.
This process optimizes two parameters: the functional parameter $\delta$
of generating function $f$ and the structural parameter $\eta$ of
graph $G$, representing a Structural Causal Model (SCM). In Phase
1 of the process, we want to keep the structural parameter $\eta$
fixed and update the functional parameter $\delta$, whereas in Phase
2, we keep $\delta$ fixed and update $\eta$. Both sets of parameters
underwent training using the buffer $B^{*}$ (Eq. 1).
The generating function $f$ is initialized as a 3-layer MLP neural
network with random parameter $\delta$. The parameter $\eta\in R^{M\times M}$,
the soft adjacency matrix of size $M\times M$ representing the direct
causality graph of the $M$ steps, is initialized as a random $M\times M$
tensor, such that $\eta_{ij}$ denotes the causal relationship between
step at index $j$ of $S_{COAS}$ on step at index $i$ of $S_{COAS}$.
At each step in lines 4 and 20 of Algo. \ref{alg:train_SCM}, we sample
a hypothesis causal graph $G$ by Bernoulli sampling $Ber((\sigma(\eta)))$
that will be used for the optimization process, where $\sigma(x)=\frac{1}{1+e^{-x}}$.

The intuition behind this optimization process is that the step representing
the \textquotedbl cause\textquotedbl{} should occur before its associated
\textquotedbl effect\textquotedbl{} step, so, for a step $t$ in
the trajectory $k$-th, we formulate $f$ as:

\begin{multline*}
(\hat{o}_{t}^{k},\hat{a}_{t}^{k})=f_{\delta,\mathcal{I}\left((o_{t}^{k},a_{t}^{k})\right)}(\{o_{t'}^{k},a_{t'}^{k}\}_{t'=1}^{t-1}\\
\wedge\text{\ensuremath{\left(\mathcal{I}(o_{t'}^{k},a_{t'}^{k})\right)}\ensuremath{\ensuremath{\in}\ensuremath{\mathrm{PA}\left(\mathcal{I}(o_{t}^{k},a_{t}^{k})\right)}}}|G)
\end{multline*}

In our implementation, every steps from $1$ to $t-1$ that do not
belong to the parental set of $\mathrm{PA}\left(\mathcal{I}(o_{t}^{k},a_{t}^{k})\right)$
are masked out when inputting into the MLP, in this way, only steps
that belong to the parental set of $\mathrm{PA}\left(\mathcal{I}(o_{t}^{k},a_{t}^{k})\right)$
are used in the prediction of $(o_{t}^{k},a_{t}^{k}).$ To learn $f$
and optimize parameter $\delta$, we compute an MSE loss as denoted
in Eq. 3. 

In the second phase, we fix $\delta$ and optimize the parameter $\eta$
by updating the causality from variable $X_{j}$ to $X_{i}$. After
updating parameter $\eta$ for several steps, we return to the optimization
process of parameter $\delta$. 

Finally, we use the resulting structural parameter $\eta$ to construct
the final causal graph $G$. We first get edge $e_{ij}$ using Eq.
4, where, $\phi_{causal}$ represents the causal
confident threshold. In our implementation, $\phi_{causal}$ was tuned
with the values in $[0.5,0.6,0.7,0.8,0.9,1.0]$. This Eq. 4
is used to ensure that there is no internal loop in the adjacency
matrix. 

\begin{algorithm}[h]
\begin{algorithmic}[1]
\label{alg:algo3}
\Require{buffer $B^{*}$, crucial step set $S_{COAS}$, mapping function $\mathcal{I}$, $\mathtt{Adam}$ optimizer, M, functional parameter ${\delta}$, structural parameter ${\eta}$,  and adjacency matrix $e$}
\State Initialize ${\delta}$, ${\eta} \in R^{M \times M}$, $e \in R^{M \times M}$
\While {$T$ times iteration}
	\For {$iteration = 1,2,...,F_s$}	
			\State $G \sim Ber((\sigma(\eta)))$		
			\For {$episode_{k}$ in $B^{*}$}
				\For {step $t=2,...,episode_{k}.length$}
				\Comment {Start from the 2nd step of the episode}
					\State $Input=\{o_{t'}^{k},a_{t'}^{k}\}_{t'=1}^{t-1}$ $\wedge$ $\left(\mathcal{I}(o_{t'}^{k},a_{t'}^{k})\right)\ensuremath{\in\mathrm{PA}\left(\mathcal{I}(o_{t}^{k},a_{t}^{k})\right)}|G$
					\State $Target = (o_{t}^{k},a_{t}^{k})$
					\State $(\hat{o}_{t}^{k},\hat{a}_{t}^{k})=f_{\delta,\mathcal{I}\left(Target\right)}(Input)$
					\State Compute $\mathcal{L}^{k}_{\delta, G}=MSE\left(Target,(\hat{o}_{t}^{k},\hat{a}_{t}^{k})\right)$
					\State $\delta \leftarrow \mathtt{Adam}(\delta, \nabla_{\delta}L_{\delta, G})$ 
				\EndFor
			\EndFor
	\EndFor
	\For {$iteration = 1,2,...,Q_s$}
			\For {$episode_{k}$ in $B^{*}$}
				\For {step $t=2,...,episode_{k}.length$}
					\State $G^{(h)} \sim Ber((\sigma(\eta)))$
					\State $Input=\{o_{t'}^{k},a_{t'}^{k}\}_{t'=1}^{t-1}$ $\wedge$ $\left(\mathcal{I}(o_{t'}^{k},a_{t'}^{k})\right)\ensuremath{\in\mathrm{PA}\left(\mathcal{I}(o_{t}^{k},a_{t}^{k})\right)}|G^{(h)}$
					\State $Target = (o_{t}^{k},a_{t}^{k})$					
					\State $(\hat{o}_{t}^{k},\hat{a}_{t}^{k})=f_{\delta,\mathcal{I}\left(Target\right)}(Input)$	
					\State Compute $\mathcal{L}^{k}_{\delta, G^{(h)}}=MSE\left(Target,(\hat{o}_{t}^{k},\hat{a}_{t}^{k})\right)$
					\State $i=S_{COAS}.index(\mathcal{I}\left(Target)\right))$
					\For {$item$ in $S_{COAS}$}
						\If {$item$ in $Input$}
							\State $j=S_{COAS}.index(\mathcal{I}\left(item)\right))$
							\State $\eta_{ij} \leftarrow \eta_{ij}-\text{\ensuremath{\beta}}(\sigma(\eta_{ij})-e_{ij}^{(h)})\mathcal{L}^{k}_{\delta,G^{(h)}}$
						\EndIf
					\EndFor
				\EndFor
			\EndFor	
	\EndFor
\EndWhile
\For {$i=1,2,...,M$} 
	\For {$j=1,2,...,M$}      
		\If{$\eta_{ij} > \eta_{ji}$ and $\sigma(\eta_{ij}) > \phi_{\text{causal}}$}
		\Comment $\phi_{causal}$ = $0.7$         
			\State $e_{ij} \gets 1$       
		\Else          
			\State $e_{ij} \gets 0$       
		\EndIf    
	\EndFor
\EndFor
\State $G$ = $\{e_{ij}\}$
\State \textbf{return} $G$
\end{algorithmic}

\caption{Train Structural Causal Model.\label{alg:train_SCM}}
\end{algorithm}

\subsubsection{Causal Tree Extraction\\}

We extract a tree from the resultant causal graph $G$, focusing on
the steps relevant to achieving the goal. We use the goal-reaching
step as the root of the tree and recursively determine the parental
steps of this root node within graph $G$ and add them to the causal
tree as new nodes, and subsequently, we will determine the parental
steps for all these identified nodes. However, to avoid cycles in
the tree, we need to add an order of ranking, thus, we use the ranking
of attention score $a_{s}$. So, for edge $e_{ij}$ from variable
$X_{i}$ to variable $X_{j}$, we will remove the edge $e_{ij}$ if
$a_{s}$ of $X_{i}$ is smaller than the $a_{s}$ of $X_{j}$, even
if $e_{ij}$=1 according to the graph $G$. 

\subsection{Setting to Test VACERL Causal Intrinsic Reward \label{subsec:Details-of-Experiments}}

\setcounter{figure}{0} 

\renewcommand{\thefigure}{\Alph{subsection}.\arabic{figure}} 

\subsubsection{Environments\label{subsec:Environments}\\}

\noindent \textit{\uline{FrozenLake Environments}}

\begin{figure}[t]
\centering{}\includegraphics[width=\linewidth]
{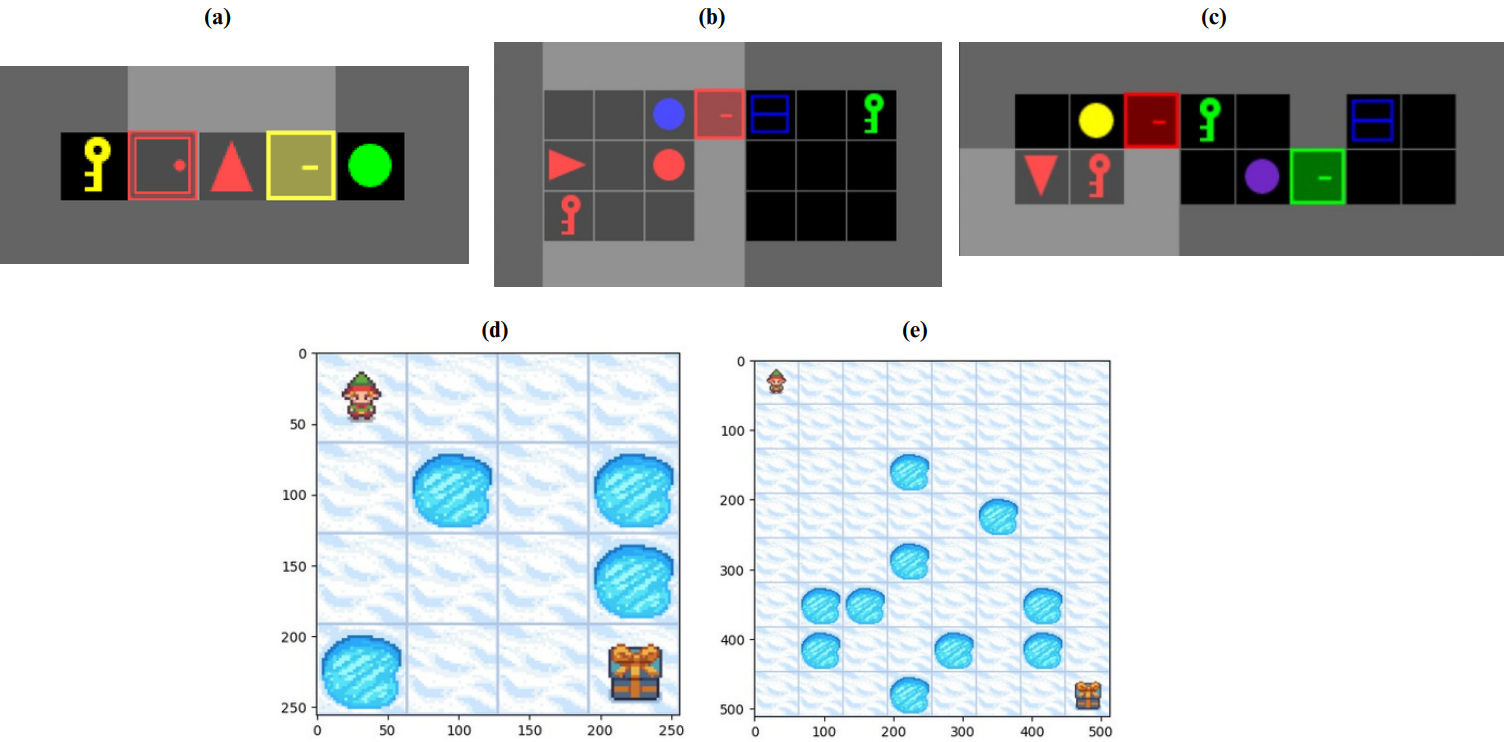}\caption{\textbf{(a)}: MG-1. \textbf{(b)}: MG-3. \textbf{(c)}: MG-4. \textbf{(d)}:
4x4FL. \textbf{(e)}: 8x8FL. \label{fig:4 envs}}
\end{figure}

These tasks involve navigating the FrozenLake environments of both
4x4 (4x4FL) and 8x8 (8x8FL) \cite{towers_gymnasium_2023}. Visualizations
for these environments can be found in Fig. \ref{fig:4 envs}(d,e).
The goal of the agent involves crossing a frozen lake from the starting
point located at the top-left corner of the map to the goal position
located at the bottom-right corner of the map without falling into
the frozen lake. The observation in these environments is a value
representing the current position of the agent. The number of possible
positions depends on the map size, with 4x4FL having 16 positions
and 8x8FL having 64 positions. The agent is equipped with four discrete
actions determining the direction of the agent's movement \cite{towers_gymnasium_2023}.
If the agent successfully reaches the goal, it receives a +1 reward.
However, if it falls into the lake or fails to reach the goal within
a predefined maximum number of steps, it receives a 0 reward. The
chosen maximum number of steps for 4x4FL and 8x8FL to validate our
framework are 100 steps and 2000 steps, respectively. 

\noindent \textit{\uline{Minihack Environments}}

These tasks involve MH-1 (MiniHack-Room-5x5-v0), MH-2 (MiniHack-Room-Monster-5x5-v0),
MH-3 (MiniHack-Room-Ultimate-5x5-v0) and MH-4 (MiniHack-River-Narrow-v0);
a suit of environments collected from \cite{samvelyan2021minihack}.
These environments present more challenging exploration scenarios
compared to FrozenLake environments due to the increased number of
objects. Certain environments necessitate interaction with objects
to achieve the goal, such as defeating monsters (MH-2 and MH-3) or
constructing bridges (MH-4). If the agent successfully reaches the
goal within a predefined maximum number of steps, it receives a +1
reward; otherwise, it receives a 0 reward. The maximum number of steps
for all four environments is the default number of steps in \cite{samvelyan2021minihack}.

\noindent \textit{\uline{Minigrid Environments}}

These tasks involve four environments: Key Corridor (MG-1 Fig. \ref{fig:4 envs}(a)),
two 2x2 rooms (MG-2 Fig. 1), two 3x3 rooms
(MG-3 Fig. \ref{fig:4 envs}(b)) and three 2x2 rooms (MG-4 Fig. \ref{fig:4 envs}(c)).
The goal of the agent, in MG-1, is to move and pick up the yellow
ball and, in MG-2, MG-3 and MG-4, is to pick up a blue box which is
located in the rightmost room, behind a locked door \cite{MinigridMiniworld23}.
In these environments, the agent has six actions: turn left (L), right
(R), move forward (F), pick up (PU), drop (D) and use (T) the object.
With each new random seed, a new map is generated, including the agent's
initial position and the position of the objects. For MG-2, MG-3,
and MG-4, there will always be an object obstructing the door. Specifically,
for MG-3 and MG-4, the agent has to pick up a key with a matching
colour to the door. In MG-3, we also introduce two distracted objects
(the red ball and the green key in Fig. \ref{fig:4 envs}(b)). All
3 environments are POMDP, meaning that the agent can only observe
part of the map; the observation is an image tensor of shape {[}7,7,3{]}.
The agent is equipped with six actions. Successfully reaching the
goal within a predefined maximum number of steps results in a +1 reward
for the agent; otherwise, it receives a 0 reward. The selected maximum
number of steps for MG-1 is 270, MG-2 is 500, MG-3 is 1000 and for
MG-4 is 5000. To complete these environments, the agent has to learn
to move the ball by picking it up and dropping it in another location,
then, it has to pick up the key, open the door and pick up the object
in the other room.

\subsubsection{Baseline Implementations \label{subsec:Detailed-Implementations-of-Intrinsic-Reward-Baseliness}\\}

The backbone algorithm is Proximal Policy Optimization (PPO). We use
the Pytorch implementation of this algorithm on the open-library Stable
Baselines 3 \footnote{{\scriptsize{}https://github.com/DLR-RM/stable-baselines3}}.
To enhance the performance of this backbone algorithm, we fine-tuned
the entropy coefficient and settled on a value of 0.001 after experimenting
with {[}0.001, 0.005, 0.01, 0.05{]}. All other parameters were maintained
as per the original repository. Subsequently, we incorporated various
intrinsic reward baselines on top of the PPO backbone, including Count-Based,
Random Network Distillation (RND), ATTENTION, CAI, and our VACERL.

For the Count-Based method, similar to the work of Bellemare et al.~\cite{bellemare2016unifying},
we tracked the frequency of observations and associated actions and
used the Simhash function to merge similar pairs \cite{tang2017exploration}.
The intrinsic reward is formulated as $r^{+}(o,a)=\frac{\alpha}{\sqrt{n(\phi(o,a))}}$,
where $n(\phi(o,a))$ represents the count and $\phi(o,a)$ is the
simhash function. We tune the exploration bonus hyperparameter $\alpha$,
however, there are no significant performance gains, as long as the
bonus rewards do not overtake the rewards of the environment. Finally,
we settle for a value of $0.001$. We also test different values of
the hashing parameters $k$ of the Simhash function. The final implementation
used $k=256$, which shows the best results and is consistent with
the referenced paper \cite{bellemare2016unifying}. 

We also adopt a public code repository for the implementation of RND
(MIT License)\footnote{{\scriptsize{}https://github.com/jcwleo/random-network-distillation-pytorch}}.
To align the implementation with our specific environments, we have
made adjustments to the input shapes and modified the PPO hyperparameters
to match those of other baselines. We adhered to the implementation
provided in the code, using the hyperparameter values as specified.

In the case of ATTENTION, we leveraged the attention scores $a_{s}$
from the encoder layer of the Transformer model as the reward signal.
This approach has been used in the work of Pitis et al., \cite{pitis2020counterfactual}
as a method to measure causal influence. The intrinsic reward has
the form $r_{bonus}\left(o,a\right)=\alpha a_{s}\left(o,a\right)$.
We integrate this as an iterative process similar to our VACERL framework
and also aid the exploration in the earlier phase using Eq. 6
for fairness. Similar to our framework VACERL and the implementation
of Count-based, we use an $\alpha$ value of $0.001$ for this baseline.

Finally, for the CAI method, we measure the causal influence between
each observation-action pair with the observation-action pair of the
goal-reaching step \cite{seitzer2021causal}. This is slightly different
from the implementation of the original paper, which assumes the knowledge
of the location of the goal and other objects. Our final implementation
used a two-layer MLP neural network to measure CAI and it is based
on the code of the original paper (MIT License)\footnote{{\scriptsize{}https://github.com/martius-lab/cid-in-rl}}). 

Additional details of hyperparameters can be found in Sec. \ref{subsec:Architecture-and-Hyperparameters}.

\subsubsection{Additional Experiment Results\label{subsec:Additional-Experiment-Results}\\}

This section presents additional experiment results and visualization. 

\noindent \textit{\uline{Learning Curve}}

\begin{figure}[t]
\begin{centering}
\centering{}\includegraphics[width=\linewidth]
{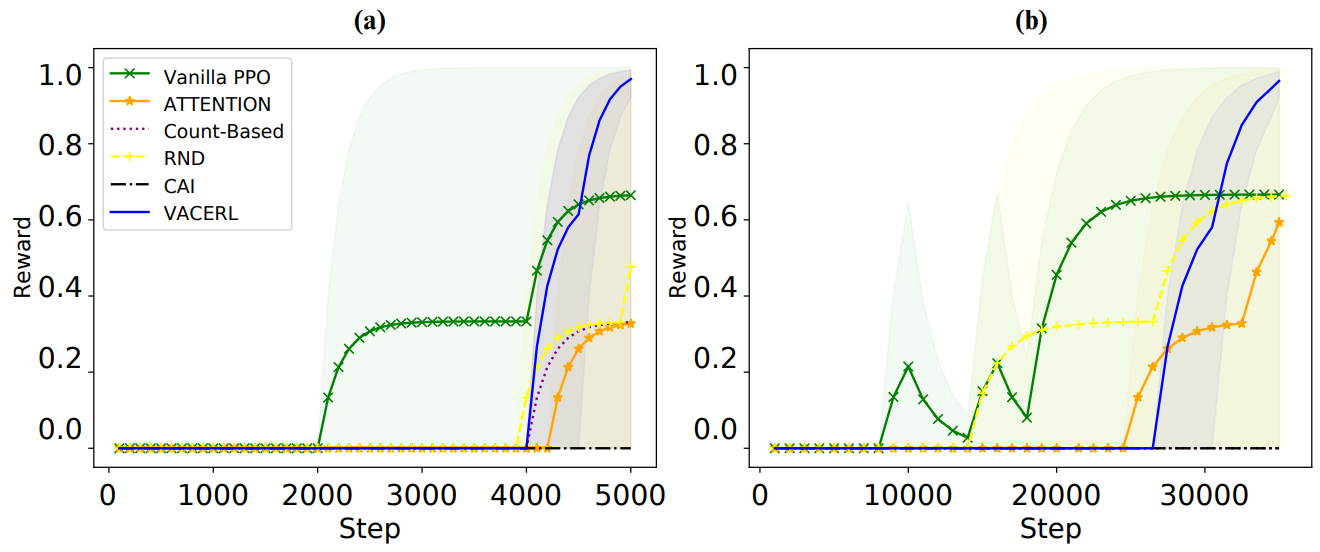}\caption{Average return over 50 episodes (mean$\pm$std over 3 runs). Learning
curve on (a) 4x4FL and (b) 8x8FL. \label{fig:Average-return-over_4x4FL_8x8FL}}
\par\end{centering}
\end{figure}

\begin{figure}[t]
\begin{centering}
\centering{}\includegraphics[width=\linewidth]
{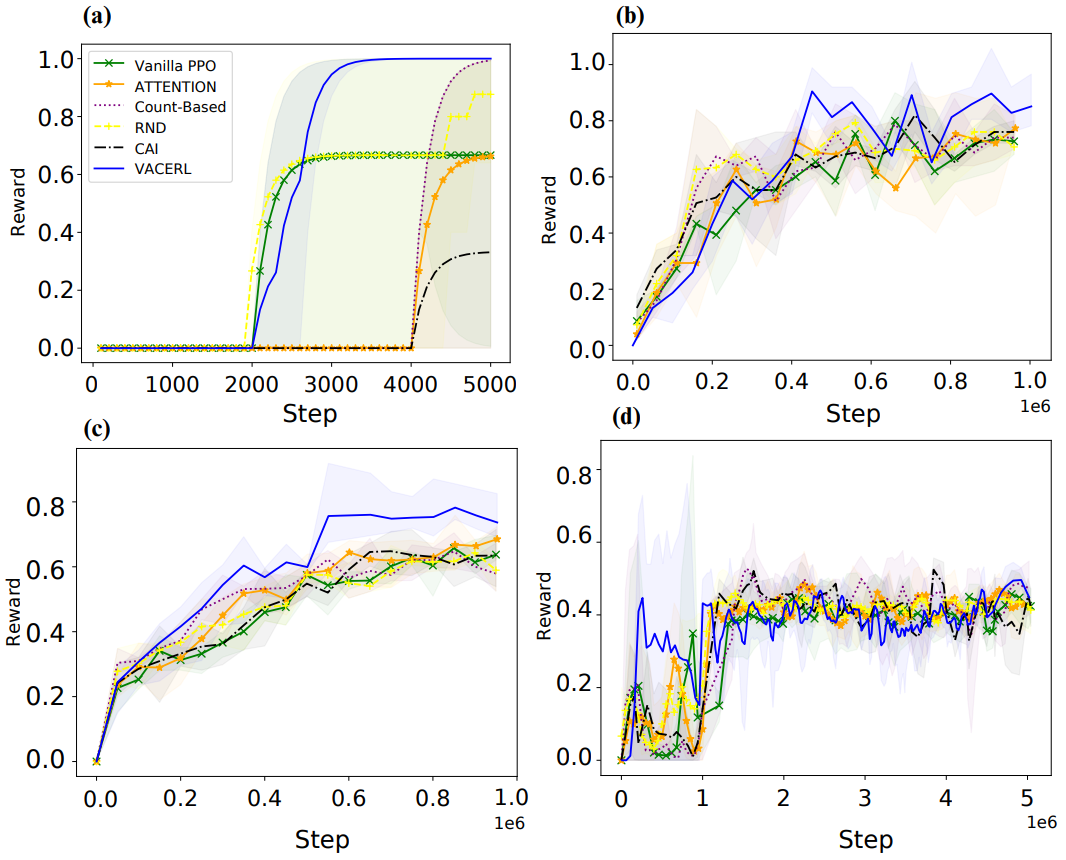}\caption{Average return over 50 episodes (mean$\pm$std over 3 runs). Learning
curve on (a) MH-1, (b) MH-2, (c) MH-3 and (d) MH-4.\label{fig:Average-return-over_MH}}
\par\end{centering}
\end{figure}

The learning curves for the 4x4FL and 8x8FL tasks are illustrated
in Fig. \ref{fig:Average-return-over_4x4FL_8x8FL}; the learning curves
for the MH-1 to MH-4 are illustrated in Fig. \ref{fig:Average-return-over_MH},
while the corresponding curves for MG-1, MG-3 and MG-4 are presented
in Fig. \ref{fig:Average-return-over_MG_2_MG_3}. The learning curves
of VACERL in these figures show a similar pattern to the learning
curves for MG-2 as presented in Fig. 2. Initially,
the learning progress is slightly slower, as a number of training
steps are required to acquire a correct causal representation. Subsequently,
the performance accelerates rapidly, eventually surpassing baselines
and attaining the optimal point. 

The steps shown in these figures are the number of times the agent
interacts with the environment, so for fairness, the number of steps
of VACERL and causal baselines (CAI and ATTENTION) are computed as
$H_{s}+a.T_{s}$, where $a$ is the number of outer-loop iterations
in line 7 of Algo. \ref{alg:VACERL-framework.}. 

\begin{figure}[t]
\begin{centering}
\includegraphics[scale=0.48]{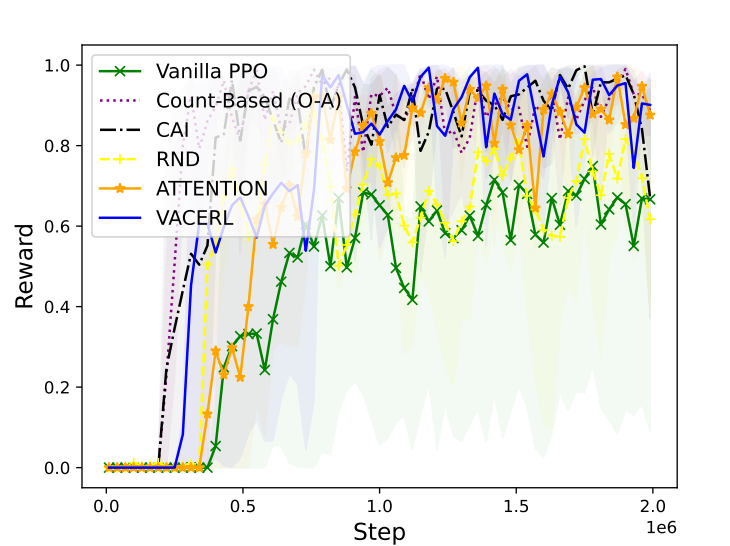}
\par\end{centering}
\caption{Average return over 50 episodes (mean$\pm$std over 3 runs). Learning
curve on MG-2 with vector input.\label{fig:Average-return-over_MG_1_vector_input}}

\end{figure}

\begin{figure}[t]
\centering{}\includegraphics[width=\linewidth]
{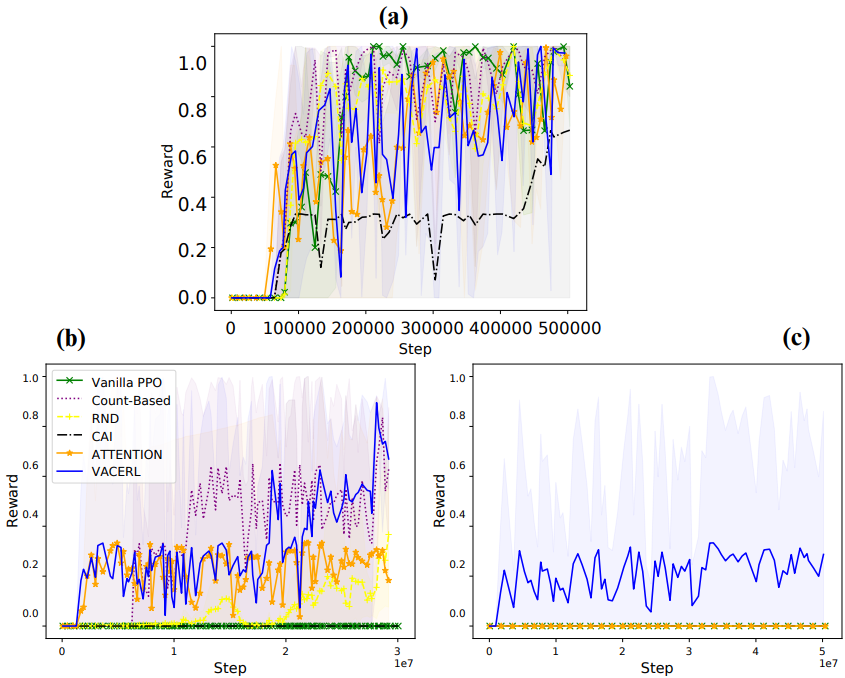}\caption{Average return over 50 episodes (mean$\pm$std over 3 runs). Learning
curve on (a) MG-1, (b) MG-3 and (c) MG-4.\label{fig:Average-return-over_MG_2_MG_3}}
\end{figure}

\noindent \textit{\uline{4x4FL Heatmap}}

The attention heatmap for the 4x4FL task is provided in Fig. \ref{fig:(a,b)-Attention-heatmap_4x4FL}.
These two figures show a similar pattern as in Fig. 4(a,b). Specifically, when the size of buffer B is small (Fig. \ref{fig:(a,b)-Attention-heatmap_4x4FL}(a)),
the accuracy of crucial step detection is not as precise compared
to scenarios where the size of buffer B is larger (Fig. \ref{fig:(a,b)-Attention-heatmap_4x4FL}(b)).

\begin{figure}[t]
\centering{}\includegraphics[width=\linewidth]
{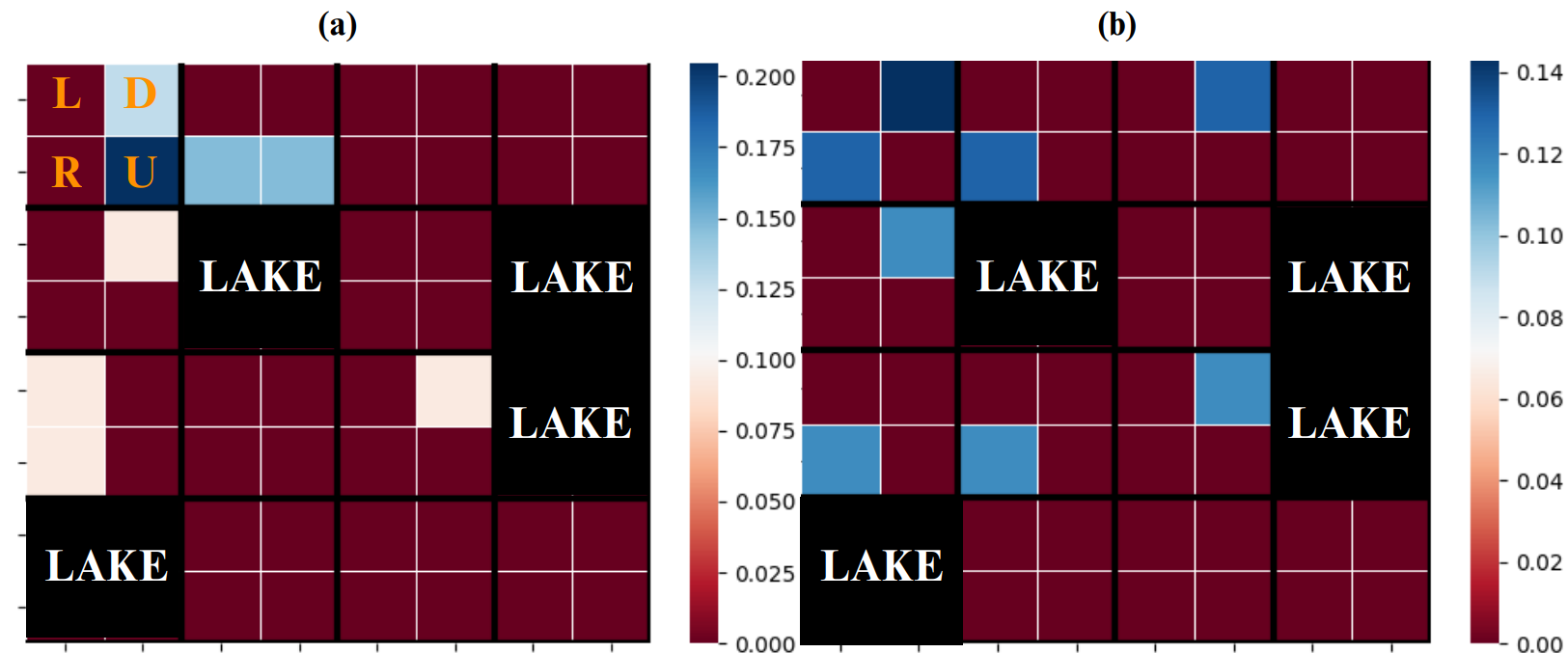}\caption{(a,b): Attention heatmap when $B$ has (a) 10 and (b) 100 trajectories
for the 4x4FL task (Fig. \ref{fig:4 envs}). We highlight the top-8
attended actions and their corresponding grids. \label{fig:(a,b)-Attention-heatmap_4x4FL}}
\end{figure}

\noindent \textit{\uline{MG-2 Generated Causal Graph}}

The causal graph generated, as a result of our framework, for the
MG-2 task is provided in Fig. 5. Each
observation-action is associated with an image representing the map
at the timestep that the agent executes the action. Below is a summary
of the relationships and our rationale for why the agent generated
the unexpected, though reasonable, relationship.
\noindent Expected relationships: 
\begin{itemize}
\item Drop Key to Pick Up Goal.
\item Open Door to Pick Up Goal.
\item Pick Up Key to Drop Key.
\item Pick Up Key to Open Door.
\item Pick Up Ball to Drop Ball.
\end{itemize}
Unexpected Relationships:
\begin{itemize}
\item Drop Ball to Pick Up Goal and not to Open Door: We believe that this
unexpected relationship is because the agent is allowed to repeat
action that it has taken before in the environment. Consequently,
the agent can pick up the ball again after it opens the door, thus
affect the relationship between Drop Ball and Open Door. In addition,
as the agent can only hold one item at a time, it must drop the ball
before picking up the goal, which creates a relationship between Drop
Ball and Pick Up Goal. If this sequence of steps frequently occurs
in collected trajectories, the agent will infer that this sequence
represents an accurate relationship, thus leading to the generation
of this causal graph. Although this relationship is not what we expected,
the relationship is not inaccurate, particularly in this environment
wherein the agent can only hold a single item at a time.
\end{itemize}

\subsection{Setting to Test VACERL with Causal Subgoal for HRL \label{subsec:Details-of-Experiments-1}\\}

\setcounter{figure}{0} 

\renewcommand{\thefigure}{\Alph{subsection}.\arabic{figure}} 

\subsubsection{Environments\label{subsec:Environments-1}\\}

Both of the environments used in this section are available in Gymnasium-Robotics
\cite{gymnasium_robotics2023github} and are built on top of the MuJoCo
simulator \cite{todorov2012mujoco}. The robot in question has 7 degrees
of freedom (DoF) and a two-fingered parallel gripper. In FetchReach,
the state space is $S\subset R^{10}$, and in FetchPickAndPlace, the
state space is $S\subset R^{25}.$ In both environments, the action
space is $A\subseteq[-1,1]^{4}$, including actions to move the gripper
and opening/closing of the gripper. In FetchReach, the task is to
move the gripper to a specific position within the robot's workspace,
which is relatively simpler compared to FetchPickAndPlace. In the
latter, the robot must grasp an object and relocate it. 

In both cases, user are given two values which are ``achieved\_goal''
and ``desired\_goal''. Here, \textquotedbl achieved\_goal\textquotedbl{}
denotes the final position of the object, and \textquotedbl desired\_goal\textquotedbl{}
is the target to be reached. In FetchReach, these goals represent
the gripper's position since the aim is to relocate it. While in FetchPickAndPlace,
they signify the block's position that the robot needs to manipulate.
Success is achieved when the Euclidean distance between \textquotedbl achieved\_goal\textquotedbl{}
and \textquotedbl desired\_goal\textquotedbl{} is less than $0.05m$. 

Sparse rewards are employed in our experiments, wherein the agent
receives a reward of $-1$ if the goal is not reached and $0$ if
it is. The maximum number of timesteps allowed in these environments
is set to 100.

\subsubsection{Baseline Implementations \label{subsec:Detailed-Implementations-of-Causality-Subgoal}\\}

We utilize the PyTorch implementation of DDPG+HER from the Stable
Baselines 3 \footnote{{\scriptsize{}https://github.com/DLR-RM/stable-baselines3}}
open-library as one of our baselines. The hyperparameters for this
algorithm are set to benchmark values available in RL-Zoo \footnote{{\scriptsize{}https://github.com/DLR-RM/rl-baselines3-zoo}}.
We assess the performance of this baseline against the results presented
in the original robotic paper by Plappert et al. \cite{plappert2018multi},
noting similarities despite differences in environment versions. 

For our HAC implementation, the core algorithm of our approach, we
adopt a publicly available code repository \footnote{{\scriptsize{}https://github.com/andrew-j-levy/Hierarchical-Actor-Critc-HAC-}}
(MIT License) by the author of the original paper \cite{levy2017learning}.
We modify this code to align with our environments, where the goal
position and the goal condition are supplied by the environments themselves.
The baseline is implemented as a three-level DDPG+HER, in which the
top two levels are used to supplied subgoals and the lowest level
is used to learn the actions. We adjust the hyperparameters of the
lowest level DDPG+HER to match those of the DDPG+HER baseline for
fairness.

Additional details of hyperparameters can be found in Sec. \ref{subsec:Architecture-and-Hyperparameters}.

\subsubsection{Additional Experiment Results \label{subsec:Additional-Experiment-Reults-1}\\}






\begin{figure}[t]
\begin{tabular}{ll}
\includegraphics[scale=0.65]{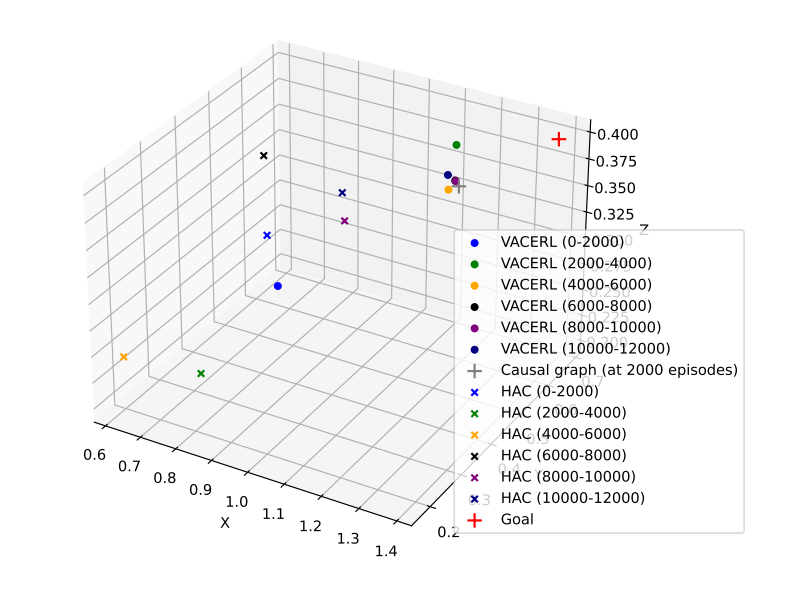}
&
\includegraphics[scale=0.28]{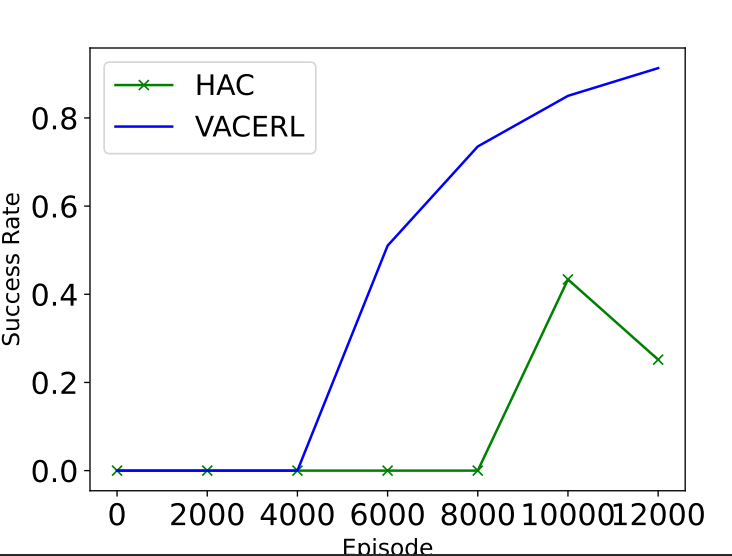}
\end{tabular}
\caption{\textbf{(a)}: The average coordinates (x, y, z) of subgoals selected
by the top level policy in VACERL (both subgoals selected randomly
and causal subgoals) and HAC at different training time intervals
(the values in parentheses indicate the start and end episodes). The
results are collected from a single run in FetchReach environment.
\textbf{(b)}: The associated learning curve of the aforementioned
run.
\label{fig:(a):-The-average}}
\label{Fig:Race}
\end{figure}

To validate our assertion that causal subgoals can effectively narrow
down the search space for an HRL agent to significant subgoals, thus
enhancing HRL sample efficiency in Robotic-Environments, we present
an additional experiment along with the visualization of subgoals'
average coordinates selected by VACERL and Vanilla HAC in this experiment
(Fig. \ref{fig:(a):-The-average}). The experiment was conducted in
the FetchReach environment, with the causal graph re-evaluated every
2,000 steps, mirroring our main experiments. We specifically chose
a run where initial subgoals of HAC and VACERL exhibited similar average
coordinates (x, y, z) for fairness. In this run, the goal (indicated
by a red + marker) was positioned at coordinates (1.38294353, 0.68003652,
0.396999). 

As illustrated in Fig. \ref{fig:(a):-The-average}(a), despite the
initial subgoals' average coordinates being very similar (represented
by blue markers) – (0.727372811, 0.501121846, 0.28868408) for HAC
and (0.7377534, 0.521795303, 0.2345436) for VACERL – VACERL swiftly
converges to subgoals much closer to the goal after just one iteration
of causal discovery learning, while, Vanilla HAC struggles to converge.
We plot the weighted average coordinates of nodes in the causal graph
after this iteration (indicated by a grey + marker), with weights
determined by the probability of node sampling according to Eq. 7;
higher probabilities correspond to higher weights. We choose to plot
the values of this iteration because it represents the instance where
VACERL undergoes the most significant shift in subgoals' coordinates.
The results indicate that the coordinates of nodes in the causal graph
closely align with the coordinates of subgoals sampled by the top-level
policy. This supports our intuition that causal subgoals contribute
to the improvement in subgoal sampling and the overall sample efficiency
of HRL.

The improvement is also reflected in the associated learning curve,
in Fig. \ref{fig:(a):-The-average}(b), of the agent: after training
for 4,000 episodes, VACERL begins to learn the environment, whereas
HAC requires 8,000 episodes – coinciding with the point where the
agent starts selecting subgoals with coordinates closer to the goal.

\subsection{Architecture and Hyperparameters of VACERL\label{subsec:Architecture-and-Hyperparameters}\\}

\setcounter{figure}{0} 

\renewcommand{\thefigure}{\Alph{subsection}.\arabic{figure}} 

The default hyperparameters (if not specified in accompanying tables
then these values were used) are provided in Table. \ref{tab:Hyperparameters-of-PPO-1}.
The definitions and values for hyperparameters, which require tuning
and may vary across different environments, are specified in the accompanying
tables. The system's architecture and the explanation for tuning of
hyperparameter are outlined below: 

\noindent \textit{\uline{Architecture}}
\begin{itemize}
\item $TF$ model's architecture: $num\_encoder\_layers=2$, $num\_decoder\_layers=2$,
$hidden\_size=128$, $dropout=0.1$.
\item Functional model $f_{\delta}$'s architecture: 3-layer MLP, $hidden\_size=512$.
\item PPO: Stable Baselines 3's hyperparameters with entropy coefficient
$=0.001.$
\item DDPG+HER: RL-Zoo's architecture and hyperparameters for FetchReach
and FetchPickAndPlace environments.
\item HAC: 3-levels DDPG+HER, architectures and hyperparameters are the
same with DDPG+HER.
\end{itemize}
\noindent \textit{\uline{Tuning}}
\begin{itemize}
\item $H_{s}$ (used for VACERL and all causal baselines): This hyperparameter
requires tuning as it relies on the complexity of the environment.
The more challenging the environment, the greater the number of head
steps required to gather a successful trajectory and start the framework.
For MG, FL, and MH environments, we use a random policy to collect
this initial phase, however, in challenging robotic environments where
collecting successful trajectories is difficult, we leverage the underlying
HAC agent to gather these trajectories. Consequently, the value of
$H_{s}$ equals $T_{s}$ in such environments. Additionally, $H_{s}$
in MG, FL, and MH denotes the number of time the agent interacts with
the environments, whereas in robotic environments, it denotes episode.
\item $M$: This hyperparameter requires tuning as it depends on the state-space
of the environment. Generally, a larger state-space requires a larger
value for $M$. However, as shown in Fig. 4, too
large $M$ can introduce noise during the causal structure discovery
phase and affect the final policy training result. 
\item $\phi_{sim}$: This hyperparameter is only used in continuous space
environments.
\item $T_{s}$ (used for VACERL and all causal baselines): Similar to $H_{s}$,
this hyperparameter varies between environments. $T_{s}$ in MG, FL,
and MH denotes the number of steps the agent interacts with the environments,
whereas in robotic environments, it denotes episode. $T_{s}$ is also
the number of steps/episodes before the causal graph is reconstructed.
\end{itemize}
\begin{table}[h]
\centering{}%
\begin{tabular}{cc||c||c||cc}
\hline 
Hyperparameters & \multicolumn{4}{c}{Def.} & Value\tabularnewline
\hline 
$lr_{TF}$ & \multicolumn{4}{c}{Learning rate of $TF$ model} & $0.001$\tabularnewline
$lr_{\delta}$ & \multicolumn{4}{c}{Learning rate of parameter $\delta$} & $0.005$\tabularnewline
$lr_{\eta}$ & \multicolumn{4}{c}{Learning rate of parameter $\eta$} & $0.005$\tabularnewline
$T$ & \multicolumn{4}{c}{No. iteration causal discovery} & $10$\tabularnewline
$F_{s}$ & \multicolumn{4}{c}{Training steps for functional parameter} & $600$\tabularnewline
$Q_{s}$ & \multicolumn{4}{c}{Training steps for structural parameter} & $600$\tabularnewline
$\phi_{causal}$ & \multicolumn{4}{c}{Causality threshold in Eq. 4} & $0.7$\tabularnewline
$\alpha$ & \multicolumn{4}{c}{Intrinsic Coef.} & $0.001$\tabularnewline
\hline 
\end{tabular}\caption{Default hyperparameters \label{tab:Hyperparameters-of-PPO-1}}
\end{table}

\begin{table}[h]
\centering{}%
\begin{tabular}{cccc}
\hline 
Hyperparameters & Def. & 4x4FL & 8x8FL\tabularnewline
\hline 
$H_{s}$ & Head steps to start the framework & $2,000$ & $15,000$\tabularnewline
$M$ & Number of top attention steps & $8$ & $16$\tabularnewline
$\phi_{sim}$ & Similary threshold in $\mathtt{is\_sim}$ function. & - & -\tabularnewline
$T_{s}$ & Training steps of policy $\ensuremath{\pi_{\theta}}$  & $2,000$ & $8,000$\tabularnewline
\hline 
\end{tabular}\caption{Hyperparameters that required tuning FL-Environments \label{tab:Hyperparameters-of-PPO}}
\end{table}

\begin{table}[h]
\centering{}%
\resizebox{\textwidth}{!}{%
\begin{tabular}{cccccc}
\hline 
Hyperparameters & Def. & MH-1 & MH-2 & MH-3 & MH-4\tabularnewline
\hline 
$H_{s}$ & Head steps to start the framework & $500$ & $10,000$ & $10,000$ & $10,000$\tabularnewline
$M$ & Number of top attention steps & $20$ & $20$ & $20$ & $20$\tabularnewline
$\phi_{sim}$ & Similary threshold in $\mathtt{is\_sim}$ function. & $0.9$ & $0.9$ & $0.9$ & $0.9$\tabularnewline
$\phi_{causal}$ & Causality threshold in Eq. 4 & $0.6$ & $0.6$ & $0.6$ & $0.6$\tabularnewline
$T_{s}$ & Training steps of policy $\ensuremath{\pi_{\theta}}$  & $2,000$ & $200,000$ & $200,000$ & $500,000$\tabularnewline
\hline 
\end{tabular}}\caption{Hyperparameters that required tuning MH-Environments\label{tab:Hyperparameters-of-PPO-2-1}}
\end{table}

\begin{table}[h]
\centering{}%
\resizebox{\textwidth}{!}{%
\begin{tabular}{cccccc}
\hline 
Hyperparameters & Def. & MG-1 & MG-2 & MG-3 & MG-4\tabularnewline
\hline 
$H_{s}$ & Head steps to start the framework & $2,000$ & $300,000$ & $800,000$ & $1,000,000$\tabularnewline
$M$ & Number of top attention steps & $20$ & $70$ & $110$ & $110$\tabularnewline
$\phi_{sim}$ & Similary threshold in $\mathtt{is\_sim}$ function. & $0.9$ & $0.9$ & $0.9$ & $0.9$\tabularnewline
$T_{s}$ & Training steps of policy $\ensuremath{\pi_{\theta}}$  & $50,000$ & $300,000$ & $1,000,000$ & $10,000,000$\tabularnewline
\hline 
\end{tabular}}\caption{Hyperparameters that required tuning MG-Environments\label{tab:Hyperparameters-of-PPO-2}}
\end{table}

\begin{table}[h]
\centering{}%
\resizebox{\textwidth}{!}{%
\begin{tabular}{cccc}
\hline 
Hyperparameters & Def. & FETCHPUSH & FETCHPICKANDPLACE\tabularnewline
\hline 
$H_{s}$ & Head episodes to start the framework & $20,000$ & $100,000$\tabularnewline
$M$ & Number of top attention steps & $20$ & $20$\tabularnewline
$\phi_{sim}$ & Similary threshold in $\mathtt{is\_sim}$ function. & $0.9$ & $0.9$\tabularnewline
$\phi_{causal}$ & Causality threshold in Eq. 4 & $0.6$ & $0.6$\tabularnewline
$T_{s}$ & Training steps of policy $\ensuremath{\pi_{\theta}}$  & $20,000$ & $100,000$\tabularnewline
\hline 
\end{tabular}}\caption{Hyperparameters that required tuning Robotic-Environments\label{tab:Hyperparameters-of-PPO-2-2}}
\end{table}